\documentclass{article}

\makeatletter
\def\input@path{{assets/}}
\makeatother

\usepackage{iclr2027_conference,times}

\usepackage{amsmath,amsfonts,bm}

\def\eqref#1{equation~\ref{#1}}

\def\1{\bm{1}}

\DeclareMathAlphabet{\mathsfit}{\encodingdefault}{\sfdefault}{m}{sl}
\SetMathAlphabet{\mathsfit}{bold}{\encodingdefault}{\sfdefault}{bx}{n}

\usepackage{booktabs,multirow,amssymb,graphicx}
\usepackage[table]{xcolor}

\definecolor{BenchPurple}{HTML}{7952B3}
\definecolor{BenchCross}{HTML}{7D8590}

\newcommand{\benchYes}{%
  \textcolor{BenchPurple}{\ensuremath{\checkmark}}}
\newcommand{\benchNo}{%
  \textcolor{BenchCross}{\ensuremath{\times}}}
\usepackage{booktabs}
\usepackage{graphicx}
\usepackage{hyperref}
\usepackage{needspace}
\usepackage{url}
\usepackage{multirow}
\usepackage{xcolor}
\usepackage{colortbl}
\usepackage{graphicx}
\usepackage{graphicx}
\usepackage{subcaption}
\usepackage{longtable}
\usepackage{wrapfig}
\usepackage{capt-of}
\usepackage{tabularx}

\definecolor{heat1}{RGB}{245,249,255} 
\definecolor{heat2}{RGB}{222,235,247} 
\definecolor{heat3}{RGB}{198,219,239} 
\definecolor{heat4}{RGB}{158,202,225} 
\definecolor{heat5}{RGB}{107,174,214} 
\definecolor{heat6}{RGB}{66,146,198}  

\definecolor{bestgreen}{RGB}{183,221,169}
\definecolor{secondgreen}{RGB}{226,239,218}

\definecolor{ResultPurple}{HTML}{7952B3}

\newcommand{\bestcell}[1]{%
  \cellcolor{ResultPurple!25}\textbf{#1}}

\newcommand{\secondcell}[1]{%
  \cellcolor{ResultPurple!10}#1}
\graphicspath{{figures/}}

\title{Diagnosing Personalization Failures in \\Mobile GUI Agents}

\author{%
\begin{minipage}[t]{\dimexpr\textwidth-2\tabcolsep\relax}
\raggedright\normalfont
{\normalsize\bfseries
\mbox{Zhixin Lin\textsuperscript{1,2}},\hspace{0.25em}
\mbox{Dongliang Xu\textsuperscript{1,*}},\hspace{0.25em}
\mbox{Jungang Li\textsuperscript{3,4}},\hspace{0.25em}
\mbox{Shidong Pan\textsuperscript{5}},\hspace{0.25em}
\mbox{Jun Feng\textsuperscript{6}},\hspace{0.25em}
\mbox{Yanbin Sun\textsuperscript{7}},\hspace{0.25em}
\mbox{Jing Li\textsuperscript{8}},\hspace{0.25em}
\mbox{Liang Liu\textsuperscript{2,\ensuremath{\dagger}}},\hspace{0.25em}
\mbox{Yue Yao\textsuperscript{1,*}}\endgraf}
\vspace{0.8em}
{\small
\textsuperscript{1}Shandong University,
\textsuperscript{2}VIVO AI Lab,
\textsuperscript{3}The Hong Kong University of Science and Technology (GZ)\\[2pt]
\textsuperscript{4}The Hong Kong University of Science and Technology,
\textsuperscript{5}Indendpendent Researcher\\[2pt]
\textsuperscript{6}Institute of Automation, Chinese Academy of Sciences,
\textsuperscript{7}Guangzhou University\\[4pt]
\textsuperscript{8}University of Technology Sydney,\\[2pt]
\textsuperscript{\ensuremath{\dagger}}Project lead.\qquad
\textsuperscript{*}Corresponding authors.\endgraf}
\end{minipage}
}

\iclrfinalcopy 

\begin{document}

\maketitle
\lhead{Preprint}
\begingroup
\renewcommand{\thefootnote}{}
\footnotetext{Email: zxlin7488@gmail.com}
\endgroup
\begingroup
\renewcommand{\thefootnote}{*}
\footnotetext{
Correspondence to: Dongliang Xu (xudongliang@sdu.edu.cn);
Yue Yao (yaoyorke@gmail.com).
}
\endgroup

\begin{abstract}
Unlike conventional GUI agents, 
personalized GUI agents require user preferences to guide a sequential execution policy over changing GUI states. 
This creates two challenges for the agentic design: 1) Determining \textit{HOW} each preference should alter the agent's execution strategy. 2) Determining \textit{WHEN} each preference should take effect along the interaction trajectory.
However, existing benchmarks primarily evaluate overall task success only, making it difficult to evaluate agents’ reasoning ability to determine \textit{HOW} and \textit{WHEN} preferences should affect execution.
To address this, we introduce \textbf{PrefDiag-Bench} with two key components: 1) a categorization of user preferences into five types, characterizing \textit{HOW} they affect agent execution; and 2) a stage-wise diagnostic framework with 1,664 probes, tracing \textit{WHEN} preferences should take effect along the interaction trajectory and where failures occur.
Experimentally, we evaluate eleven GUI agent systems, including end-to-end agents and agentic workflows, under two forms of user context: explicit user profiles and behavioral memory logs.
We find while existing GUI agents achieve relatively high task completion rates on non-personalized tasks, their performance remains substantially lower on preference-personalized tasks. 
Even the best-performing model, Seed-2.0-Pro, achieves only 47.17\% on process preferences under explicit user profiles. Moreover, agents struggle particularly with Ranking and Process preferences, indicating substantial room for improvement. 
Further analysis reveals two key capability bottlenecks behind these failures: selecting relevant evidence from user history, and recognizing when user preferences should take effect during execution. Controlled interventions that explicitly provide relevant evidence improve end-to-end task success by up to 7.06\%. 
We hope PrefDiag-Bench can provide a more systematic and comprehensive evaluation of GUI agents’ personalization capabilities, moving beyond simply measuring overall task success. All code and dataset are available at \underline{\url{https://github.com/Zhixin-L/PrefDiag-Bench}}.

\end{abstract}

\section{Introduction}
\label{sec:introduction}
Recent mobile GUI agents powered by large language models (LLMs) have made substantial progress in GUI perception, task planning, and task execution in real-world applications, bringing them closer to personalized assistants. In such a personalization setting, simply completing a user request is no longer sufficient; agents also are required to draw on users’ past behavior, habits, and preferences to make decisions that meet individual needs during execution. 
For example, for the same base task ``order a milk tea according to my usual preferences and add it to the cart'' the agent may need to select the preferred drink options, satisfy a price constraint, prioritize a store with faster delivery, or compare multiple candidates. Therefore, the same GUI task may impose different execution requirements under different user preferences.

\begin{figure}[t]
    \centering
    \includegraphics[width=0.85\textwidth]{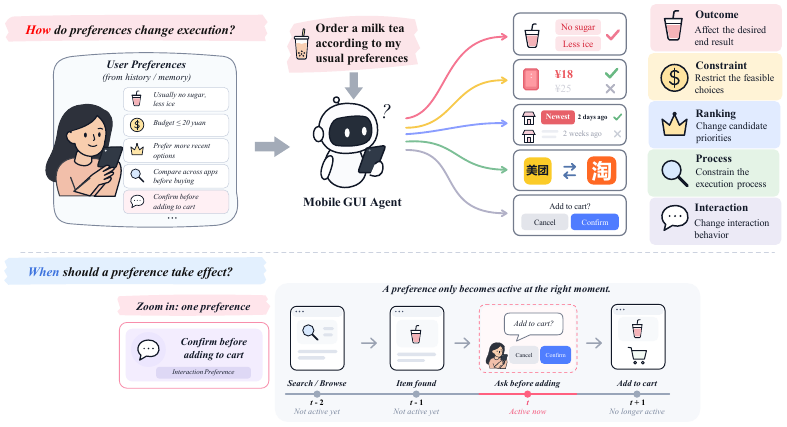}
    \caption{
    \textbf{Motivation for PrefDiag-Bench.}
    \textit{Top:} Different user preferences impose distinct execution
    requirements on mobile GUI agents, affecting desired outcomes,
    feasible choices, candidate priorities, execution procedures,
    and user interaction.
    \textit{Bottom:} The interaction example illustrates that knowing
    a task-relevant preference is insufficient: the agent must also
    recognize the appropriate GUI state in which to apply it.
    PrefDiag-Bench complements end-to-end evaluation with stage-wise
    diagnostic probes across information processing, contextual reasoning,
    and action formulation to identify capability gaps in preference
    application.
    }
    \label{fig:motivation}
    \vspace{-4pt}
\end{figure}

The example above highlights two challenges for personalized GUI agents. First, an agent is required to determine \textit{HOW} each preference should alter execution: it may change the desired outcome, restrict valid choices, change candidate priorities, or constrain the execution process or interaction behavior. Second, the agent is required to determine \textit{WHEN} a preference should take effect along the interaction trajectory, based on the current task and GUI state. Figure~\ref{fig:motivation} illustrates these two complementary challenges: preferences impose distinct execution requirements, while a task-relevant preference should only be applied when its trigger state is reached. These challenges require evaluation beyond task completion. A benchmark should assess whether an agent can determine \textit{HOW} a preference should alter execution and \textit{WHEN} it should take effect, while also identifying where failures arise along the path from user context to GUI actions. Existing evaluations centered on task-level success or preference-sensitive progress provide limited information about this process.

To address these challenges from a data-centric perspective, we introduce PrefDiag-Bench, a fine-grained diagnostic benchmark for personalized GUI agents. PrefDiag-Bench contains two complementary components corresponding to the challenges above. First, while the benchmark spans diverse application domains, including shopping, workflow, and transportation, its main evaluation axis is the primary effect of each preference on agent execution. We organize user preferences into five execution-oriented types, \emph{i.e.}, Outcome, Constraint, Ranking, Process, and Interaction—characterizing \textit{HOW} they affect agent execution, and construct corresponding end-to-end tasks to compare agents under different execution requirements. Second, we develop a stage-wise diagnostic framework that covers three high-level phases, \emph{i.e.}, information processing, contextual reasoning, and action formulation, and further divides them into five diagnostic stages. The framework probes the capabilities required to apply preferences from user context to GUI actions, including recognizing \textit{WHEN} a preference should take effect, and identifies capability bottlenecks arise during personalized execution.
Based on this design, PrefDiag-Bench contains 318 end-to-end GUI tasks and 1,664 associated diagnostic probes. 
Each end-to-end task is paired with a set of diagnostic probes built around the same underlying preference instance. This paired design allows us not only to evaluate whether an agent completes the task while satisfying the user’s preference in an interactive GUI environment, but also to connect end-to-end performance with specific diagnostic stages. In addition, we construct explicit user profiles and behavioral memory logs as two user-context representations at different levels of abstraction, enabling us to study how the form of user context affects personalized execution.

Experimentally, we systematically evaluate 11 existing popular personalized GUI agentic designs, including nine end-to-end agents and two agentic workflows, M3A and T3A, using explicit user profiles and behavioral memory logs. 
Our results indicate three main findings. First, performance varies substantially across the five preference types, showing that collapsing performance across preference types into a single aggregate score can obscure important capability differences. Second, explicit user profiles do not consistently outperform behavioral memory logs, suggesting that compact preference summaries are not always more useful than logs that retain concrete behavioral evidence. Third, selecting relevant evidence from user history and recognizing when preferences should take effect emerge as the two main bottlenecks. Controlled interventions on a representative subset of systems further show that explicit evidence guidance improves end-to-end task success by up to 7.06 percentage points, demonstrating that evidence selection has a measurable impact on personalized execution and provides a concrete target for system improvement.




\section{Related Work}
\label{sec:related-work}


\subsection{Mobile GUI Agents and Their Personalized Evaluation}

Recent mobile GUI agents have made substantial progress in GUI perception, task planning, and multi-step task execution~\citep{liu2025llmpoweredguiagentsphone,nguyen-etal-2025-gui,hu-etal-2025-os}. Existing systems include agent frameworks that organize task execution through explicit reasoning and control modules, such as AppAgent~\citep{ZhangYLLHCHF025}, M3A~\citep{ICLR2025_01a83bc2}, and Mobile-Agent-v3~\citep{ye2025mobileagentv3fundamentalagentsgui}, as well as end-to-end agents that directly generate reasoning traces and actions from GUI observations, such as UI-TARS~\citep{qin2025ui} and GUI-Owl~\citep{ye2025mobileagentv3fundamentalagentsgui}. As the GUI capabilities of these systems continue to improve, recent studies have begun to examine how agents can use user history and preferences to adapt their behavior during task execution. PersonalAlign~\citep{lyu-etal-2026-personalalign} uses long-term user records to recover preferences omitted from vague instructions; PSPA-Bench~\citep{nie2026pspabenchpersonalizedbenchmarksmartphone} decomposes tasks to evaluate preference-sensitive progress; KnowU-Bench~\citep{chen2026knowubenchinteractiveproactivepersonalized} uses behavioral logs to assess preference inference, user clarification, and proactive decision-making; and Persona2Web~\citep{kim2026persona2webbenchmarkingpersonalizedweb} studies contextual reasoning from long-term user histories in web environments. These studies broaden personalized agent evaluation from complementary directions. However, they do not jointly examine how different preferences affect GUI execution or which capabilities become bottlenecks when preferences are applied from user context to GUI actions. PrefDiag-Bench complements this line of work with an execution-oriented and stage-wise diagnostic perspective.

\subsection{Personalization User Context and Memory}
Personalization relies on memory as a key source of user context, allowing agents to retain and reuse information from past interactions~\citep{zhang2025tmlr-personalization,liu2025surveypersonalizedlargelanguage}. Memory systems such as MemoryBank~\citep{DBLP:conf/aaai/ZhongGGYW24} and Mem0~\citep{ChhikaraKASY25} extract, update, and retrieve user information from ongoing interactions to support long-term personalization. Corresponding benchmarks examine this process from different perspectives. PersonaMem~\citep{JHCLYCea25} evaluates whether models can track evolving user profiles from multi-session interaction histories and generate personalized responses accordingly. AlpsBench~\citep{Xiao_2026} decomposes personalized memory management into information extraction, updating, retrieval, and utilization. In mobile GUI environments, MemGUI-Bench~\citep{liu2026memguibenchbenchmarkingmemorymobile} focuses on memory retention and use across steps, applications, and sessions.However, correctly inferring or retrieving user information does not necessarily mean that an agent can apply user preferences correctly in a GUI task. Personalized GUI agents must further connect user context with the current task and GUI state, and translate relevant preferences into concrete behaviors such as candidate selection, ranking, process control, or user interaction. PrefDiag-Bench complements these studies by comparing explicit user profiles and behavioral memory logs in GUI tasks and probing capability gaps from user-context processing to action formulation.

\section{Benchmark Design}
\label{sec:benchmark}

PrefDiag-Bench evaluates personalized mobile GUI agents at both the task and diagnostic levels. It contains 318 end-to-end GUI tasks, including 53 Basic and 265 preference-conditioned instances and 1,664 diagnostic probes, where each end-to-end task is linked to probes constructed from the same underlying preference instance.
\begin{table*}[t]
\centering
\caption{
\textbf{Comparison with existing personalization benchmarks.}
EUP and BML denote explicit user profiles and behavioral memory logs,
respectively.
Personalization reasoning explicitly evaluates \textit{how} preferences
alter execution requirements and \textit{when} they take effect as
complementary dimensions.
Specific diagnosis pairs individual task instances with dedicated
capability probes.
}
\label{tab:benchmark_comparison}
\renewcommand{\arraystretch}{1.2}
\setlength{\tabcolsep}{7pt}

\resizebox{0.95\textwidth}{!}{%
\begin{tabular}{lcccccc}
\toprule
\multirow{2}{*}[-0.35ex]{\textbf{Benchmark}}
& \multirow{2}{*}[-0.35ex]{\textbf{Environment}}
& \multicolumn{2}{c}{\textbf{User Context}}
& \multirow{2}{*}[-0.35ex]{%
    \shortstack{\textbf{Personalization}\\\textbf{Reasoning}}}
& \multirow{2}{*}[-0.35ex]{%
    \shortstack{\textbf{E2E GUI}\\\textbf{Tasks}}}
& \multirow{2}{*}[-0.35ex]{%
    \shortstack{\textbf{Specific}\\\textbf{Diagnosis}}} \\
\cmidrule(lr){3-4}
& & \textbf{EUP} & \textbf{BML} & & & \\
\midrule

AlpsBench~\citep{Xiao_2026}
& Dialogue
& \benchYes & \benchYes
& \benchNo & \benchNo & \benchNo \\

AMemGym~\citep{ICLR2026_0856bc55}
& Dialogue
& \benchNo & \benchYes
& \benchNo & \benchNo & \benchYes \\

\addlinespace[4pt]

Persona2Web~\citep{kim2026persona2webbenchmarkingpersonalizedweb}
& Web
& \benchYes & \benchYes
& \benchNo & \benchYes & \benchNo \\

\addlinespace[4pt]

PSPA-Bench~\citep{nie2026pspabenchpersonalizedbenchmarksmartphone}
& Mobile
& \benchYes & \benchNo
& \benchNo & \benchYes & \benchNo \\

PersonalAlign~\citep{lyu-etal-2026-personalalign}
& Mobile
& \benchNo & \benchYes
& \benchNo & \benchNo & \benchNo \\

KnowU-Bench~\citep{chen2026knowubenchinteractiveproactivepersonalized}
& Mobile
& \benchNo & \benchYes
& \benchNo & \benchYes & \benchNo \\

\midrule
\rowcolor{BenchPurple!8}
\textbf{PrefDiag-Bench}
& \textbf{Mobile}
& \benchYes & \benchYes
& \benchYes & \benchYes & \benchYes \\

\bottomrule
\end{tabular}%
}
\end{table*}

\subsection{Preference Taxonomy}
\label{sec:preference-taxonomy}

User preferences concern different application domains and semantic
content. For mobile GUI agents, however, a more direct distinction
is the type of decision that each preference changes during task
execution. We therefore use preferences' primary effects on execution
as our main evaluation axis, classifying each preference by the part
of a personalized task's success criterion that it directly changes.
This yields five types, illustrated in Figure~\ref{fig:motivation}.

Specifically, as shown in the Figure~\ref{fig:motivation}, \textbf{Outcome} preference specifies the desired outcome or final
choice, such as selecting ``no sugar'' and ``less ice'' in a
drink-ordering task.
A \textbf{Constraint} preference sets conditions that a valid outcome
or allowed action must satisfy, such as keeping the drink price at
or below 20\$.
A \textbf{Ranking} preference specifies priorities among valid
candidates, rather than determining whether a candidate satisfies
a hard constraint.
For example, when several stores offer the target drink, the agent
should prioritize the store with the shortest delivery time and
choose the cheaper store when delivery times are equal.
A \textbf{Process} preference specifies how a task should be carried
out, such as comparing the price of the same product across multiple
platforms before making a choice.
The final GUI state alone is insufficient to verify this requirement:
the execution trajectory must also show that the prescribed procedure
was followed.
An \textbf{Interaction} preference specifies whether, when, and how
the agent should interact with the user.
For example, after finding a suitable item, the agent should request
confirmation and add it to the cart only after the user agrees.
Although both types affect the interaction trajectory, Process
constrains the task procedure, whereas Interaction constrains
communication with the user.
A task may therefore reach the expected final state yet still violate
its preference requirements if a required comparison or confirmation
was skipped during execution.
\subsection{Benchmark Overview and Statistics}
\label{sec:benchmark-overview}

\begin{figure}[t]
    \centering
    \includegraphics[width=\textwidth]{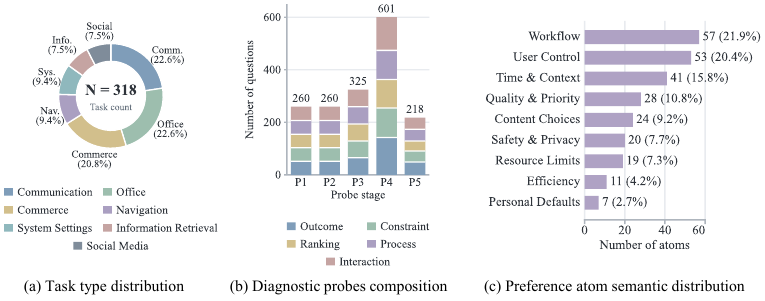}
    \caption{
    Statistics of PrefDiag-Bench.
    (a) Composition of end-to-end tasks across application domains.
    (b) Distribution of diagnostic probes across the five diagnostic stages.
    (c) Distribution of fine-grained preference atoms, showing the semantic
    diversity of user preferences covered by the benchmark.
    }
    \label{fig:benchmark-statistics}
\end{figure}

PrefDiag-Bench evaluates agents' ability to apply user preferences
across diverse mobile tasks.
It comprises 53 task families, each with one \textit{Basic} task
and five preference-conditioned variants, yielding 318 instances
(53 \textit{Basic} and 265 personalized).
As shown in Figure~\ref{fig:benchmark-statistics}(a), the tasks span
seven application domains: communication, office, commerce,
navigation, system settings, information retrieval, and social media.
Within the diagnostic set, preference atoms also cover nine semantic
categories, including workflow, user control, time and context,
resource limits, and safety and privacy
(Figure~\ref{fig:benchmark-statistics}(c)).
These categories describe what preferences concern, complementing
the taxonomy that distinguishes how they alter
agent behavior.

Formally, we define a personalized instance as
$\mathcal{T}_i=(I_i,C_i,E_i^0,S_i)$,
where $I_i$ is the instruction, $C_i$ is the user context conveying
preference $p_i$, $E_i^0$ is the initial environment state, and $S_i$
is a deterministic success checker.
The agent is required to identify the relevant preference requirements from $C_i$
and complete the task while satisfying all of them at the appropriate
stages of execution.
Execution produces a trajectory $\tau_i=(o_1,a_1,\ldots,o_T,a_T)$
of observations and actions.
The checker evaluates the final state $E_i^T$ and relevant action
or user--agent interaction records, returning
$S_i(E_i^T,\tau_i)\in\{0,1\}$.
A result of 1 requires satisfying both the task goal and all
preference-specific requirements, including any prescribed procedure
or confirmation.

To examine the capabilities underlying these task outcomes, we include
1,664 diagnostic probes associated with 260 preference atoms from
256 tasks with observed failures under behavioral logs
(Figure~\ref{fig:benchmark-statistics}(b)).
Each atom is the smallest preference requirement that can be
diagnosed and applied independently.
Across information processing, contextual reasoning, and action
formulation, the probes test preference induction from logs (P1),
evidence selection for a given preference (P2), task relevance (P3),
state-level trigger recognition (P4), and semantic action selection
for an active preference (P5).
They are evaluated independently, with ground-truth prerequisites
supplied when needed; 
This pairing connects task outcomes with tests of specific
capabilities rather than describing personalization only through
aggregate task success.

\subsection{Data Construction and Annotation}
\label{sec:data-construction}

We construct preference-conditioned variants from selected KnowU-Bench
tasks~\citep{chen2026knowubenchinteractiveproactivepersonalized},
inspecting their goals, application settings, and success criteria.
Every resulting variant is manually reviewed.
The target preference is conveyed through user context rather than
restated in the instruction.
We construct paired contexts by first drafting explicit user profiles
with an LLM, revising and checking them with GPT-5.6 Sol, and
reviewing every profile manually.
Synthetic behavioral memory logs are then generated from the reviewed
profiles, with four to five profile-related entries per collection,
and undergo the same model-assisted revision and human review.
The underlying preference, instruction, initial environment state,
and success checker remain fixed across the two context forms.

We derive task-linked diagnostic probes from execution failures under
behavioral logs, retaining cases with valid checker results and
deduplicating them across agents by task.
Independently diagnosable preference requirements are separated into
atoms.
Probe questions, options, and reference annotations are drafted with
LLM assistance and revised by humans.
For P4 and P5, we use atoms with clear trigger boundaries and usable
reference trajectories.
P5 reference actions are extracted from successful rollouts of the
corresponding tasks and represented as semantic operations.
Human reviewers answer all multiple-choice probes.

\section{Experiments}
\label{sec:experiments}
\begin{table*}[t]
\centering
\caption{
End-to-end Task Success Rates (TSR, \%) on PrefDiag-Bench.
\textit{Basic} reports traditional non-personalized task success rates averaged over the 53 base tasks.
The remaining columns report personalization task success rates across five preference types under
explicit user profiles and behavioral memory logs, together with their averages.
Constr.\ and Interact.\ denote Constraint and Interaction, respectively.
Within each category, systems are sorted in ascending order by the mean
of the two AVG scores.
The \protect\colorbox{violet!20}{\textbf{best}} and
\protect\colorbox{violet!8}{second-best} results in each column
are highlighted across all systems.
}
\label{tab:e2e-results}

\small
\setlength{\tabcolsep}{3.2pt}
\renewcommand{\arraystretch}{1.18}

\resizebox{0.95\textwidth}{!}{%
\begin{tabular}{@{}l c *{5}{c} c @{\hspace{10pt}} *{5}{c} c@{}}
\toprule

\multirow{2}{*}{\textbf{System}}
& \multirow{2}{*}{\textbf{Basic}}
& \multicolumn{6}{c}{\textbf{Explicit User Profiles}}
& \multicolumn{6}{c}{\textbf{Behavioral Memory Logs}} \\

\cmidrule(lr){3-8}
\cmidrule(l){9-14}

& & Outcome & Constr. & Ranking & Process & Interact. & \textbf{AVG}
  & Outcome & Constr. & Ranking & Process & Interact. & \textbf{AVG} \\

\midrule

\multicolumn{14}{c}{%
    \cellcolor{black!7}\strut\textit{Agentic workflows}%
} \\
\addlinespace[2pt]

T3A
& 41.51
& 37.74 & 30.19 & 13.21 & 15.09 & 28.30 & 24.91
& 43.40 & 30.19 & 22.64 & 18.87 & 32.08 & 29.44 \\

M3A
& 54.72
& 47.17 & 50.94 & 33.96 & 16.98 & 39.62 & 37.73
& 28.30 & 20.75 & 16.98 & 9.43 & 16.98 & 18.49 \\

\midrule

\multicolumn{14}{c}{%
    \cellcolor{black!7}\strut\textit{End-to-end agents}%
} \\
\addlinespace[2pt]

Qwen3-VL-8B-Thinking
& 30.19
& 20.75 & 11.32 & 13.21 & 3.77 & 9.43 & 11.70
& 22.64 & 20.75 & 11.32 & 3.77 & 1.89 & 12.07 \\

Step-GUI-4B
& 37.74
& 33.96 & 13.21 & 15.09 & 3.77 & 0.00 & 13.21
& 30.19 & 13.21 & 16.98 & 1.89 & 0.00 & 12.45 \\

Qwen3-VL-8B-Instruct
& 35.85
& 28.30 & 13.21 & 16.98 & 13.21 & 3.77 & 15.09
& 33.96 & 16.98 & 15.09 & 7.55 & 16.98 & 18.11 \\

MAI-UI-8B
& 58.49
& 45.28 & 26.42 & 24.53 & 13.21 & 11.32 & 24.15
& 56.60 & 18.87 & 22.64 & 13.21 & 13.21 & 24.91 \\

GUI-Owl-1.5-8B-Instruct
& 56.60
& 47.17 & 32.08 & 18.87 & 16.98 & 22.64 & 27.55
& 50.94 & 15.09 & 26.42 & 15.09 & 15.09 & 24.53 \\

UI-Venus-1.5-8B
& 52.83
& 39.62 & 35.85 & 24.53 & 13.21 & 26.42 & 27.93
& 47.17 & 28.30 & 28.30 & 20.75 & 22.64 & 29.43 \\

Qwen3.7-Plus
& 67.92
& 56.60 & 56.60 & 41.51 & 22.64 & \secondcell{45.28} & 44.53
& 62.26 & 49.06 & 33.96 & 16.98 & \secondcell{47.17} & 41.89 \\

Claude Sonnet 4.6
& \bestcell{86.79}
& \secondcell{71.70}
& \bestcell{69.81}
& \bestcell{58.49}
& \secondcell{35.85}
& 41.51
& \secondcell{55.47}
& \secondcell{75.47}
& \secondcell{60.38}
& \bestcell{47.17}
& \secondcell{39.62}
& 45.28
& \secondcell{53.58} \\

Seed-2.0-Pro
& \secondcell{81.13}
& \bestcell{79.25}
& \secondcell{66.04}
& \secondcell{56.60}
& \bestcell{47.17}
& \bestcell{64.15}
& \bestcell{62.64}
& \bestcell{83.02}
& \bestcell{62.26}
& \secondcell{45.28}
& \bestcell{60.38}
& \bestcell{77.36}
& \bestcell{65.66} \\

\bottomrule
\end{tabular}%
}
\end{table*}

We organize our experiments around the following five research questions.

\noindent\textbf{RQ1.}
How well do GUI agents satisfy different preference requirements
during task execution?

\noindent\textbf{RQ2.}
What strengths and weaknesses do agents exhibit across stages
of preference application?

\noindent\textbf{RQ3.}
What relationships emerge between task-level diagnostic results
and end-to-end outcomes?

\noindent\textbf{RQ4.}
What effect does evidence guidance have on end-to-end task success?

\noindent\textbf{RQ5.}
What changes in performance occur under different
memory construction methods?
\subsection{Experimental Setup}
\paragraph{Evaluated Systems.}
We evaluate 11 systems.
The closed-source agents are
Seed-2.0-Pro~\citep{seed2026seed20modelcardintelligence},
Claude Sonnet 4.6~\citep{anthropic2026sonnet46}, and
Qwen3.7-Plus~\citep{alibaba2026qwen37plus}.
The open-source agents are
MAI-UI-8B~\citep{zhou2025maiuitechnicalreportrealworld},
Qwen3-VL-8B-Instruct and
Qwen3-VL-8B-Thinking~\citep{abs-2511-21631},
Step-GUI-4B~\citep{abs-2512-15431},
GUI-Owl-1.5-8B-Instruct~\citep{xu2026mobileagentv35multiplatformfundamentalgui},
and UI-Venus-1.5-8B~\citep{venusteam2026uivenus15technicalreport}.
We also evaluate two agentic workflows,
M3A and T3A~\citep{ICLR2025_01a83bc2}, both using GPT-5.4.
Open-source models run on a server with eight NVIDIA RTX 4090 GPUs.

\paragraph{Evaluation Protocol.}
All 11 systems are evaluated on 53 \textit{Basic} tasks and
265 preference-conditioned tasks, the latter under both explicit
user profiles and behavioral memory logs.
All systems share task environments and checkers while retaining
their native observation and action interfaces.
Instructions, initial states, and checkers remain fixed across
profile--log pairs.
We report Task Success Rate (TSR) by preference type and context form.
Independent P1--P5 probes are applied to the nine directly queried
agents; workflows are excluded to avoid assessing only their backbone
or introducing workflow-specific probe adaptations.
Diagnostic results are aggregated by task, preference, and probe type:
a result is correct only if all associated items are answered correctly.
We report accuracy per probe type and an overall score across all
task-level diagnostic results.

\subsection{End-to-End Results}
\label{sec:end-to-end-results}

Table~\ref{tab:e2e-results} reports TSR on the 53 \textit{Basic} tasks and the 265 preference-conditioned tasks.
For each personalized task, we compare explicit user profiles with behavioral memory logs while keeping the task instruction, initial environment state, and success checker fixed.
The AVG columns report the macro-average over the five preference types.

\paragraph{Personalized tasks are more challenging to execute.}
All 11 systems achieve lower average TSR on preference-conditioned tasks than on \textit{Basic} tasks under both forms of user context.
Seed-2.0-Pro obtains the strongest personalized results, reaching 62.64\% with explicit profiles and 65.66\% with behavioral logs, compared with 81.13\% on \textit{Basic}.
In contrast, Claude Sonnet 4.6 achieves the highest \textit{Basic} TSR of 86.79\%, but is surpassed by Seed-2.0-Pro under both personalized settings.
This ranking change shows that strong general GUI execution is important, but is not sufficient for applying user preferences throughout a multi-step interaction trajectory.
\begin{figure}[t]
    \centering
    \includegraphics[width=\textwidth]{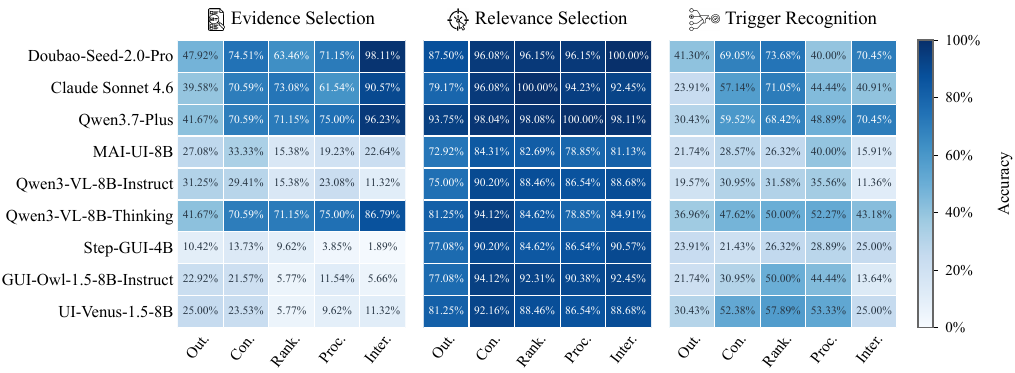}
    \caption{
    The heatmaps show P2 Evidence Selection, P3 Relevance Selection, and
P4 Trigger Recognition. Rows represent evaluated systems and columns
represent the five preference types. Darker colors indicate higher
task-level accuracy.
    }
    \label{fig:probe-heatmap}
    \vspace{-2pt}
\end{figure}
\paragraph{Different preference types expose distinct performance patterns.}
Outcome is the best-performing type for 10 of the 11 systems under
explicit user profiles and for all 11 systems under behavioral memory
logs.
At the other end, Ranking and Process emerge as recurring difficulties
across systems and user-context forms.
Process is the lowest-performing type for seven systems under explicit
profiles and the lowest or tied-lowest type for eight systems under
behavioral memory logs.
Under behavioral memory logs, Qwen3.7-Plus drops from 62.26\%
on Outcome to 16.98\% on Process, a gap of 45.28 percentage
points.
Ranking remains difficult even for the strongest system:
Seed-2.0-Pro achieves 83.02\% on Outcome but only 45.28\%
on Ranking under behavioral memory logs, a difference of 37.74
points.
These results show that reporting only an overall personalized score
would hide major differences in the execution requirements that each
system can handle.
The repeated difficulty on Ranking and Process is consistent with the
additional candidate comparison required by Ranking and the longer
execution procedure required by Process.

\paragraph{More explicit user context is not always more effective.}
The profile--log comparison has no consistent direction across systems.
Behavioral logs improve Seed-2.0-Pro from 62.64\% to 65.66\% and T3A from 24.91\% to 29.44\%, whereas M3A drops from 37.73\% with profiles to 18.49\% with logs.
This contrast is notable because M3A and T3A use the same GPT-5.4 backbone, suggesting that the usefulness of a context form also depends on how the overall workflow processes user information.
This pattern is consistent with the trade-off built into the two representations: profiles directly provide compact preference rules, whereas logs preserve concrete behavioral evidence but require the agent to identify which evidence is relevant.
These task-level comparisons reveal performance gaps across execution
requirements and context forms, but do not isolate the capabilities
underlying preference application.
We next examine these capabilities using independent probes
constructed from failures under behavioral memory logs.

\subsection{Stage-Wise Diagnostic Results}

Figure~\ref{fig:probe-heatmap} reports task-level accuracy
on P2--P4 for the nine directly evaluated agents. P1 and P5 are omitted from the heatmaps because they are near ceiling for most systems, but are included in the aggregate statistics below.

\paragraph{Evidence selection and trigger recognition are the weakest measured stages.}
On the failure-derived diagnostic set, P2 Evidence Selection and
P4 Trigger Recognition are the lowest-scoring stages across systems,
averaging 40.06\% and 39.69\%, respectively, compared with
88.80\% for P3 and over 98.9\% for P1 and P5.
Even Seed-2.0-Pro reaches only 71.48\% on P2 and 58.14\%
on P4.
Since both probes provide the target preference, these results
highlight the need to improve agents' ability to identify supporting
evidence in user history and recognize when a known preference
should take effect.

\paragraph{The Thinking variant stands out among open-source agents.}
Qwen3-VL-8B-Thinking achieves the highest accuracy among the
evaluated open-source agents on both P2 and P4.
On P2, it scores 69.53\%, compared with 21.88\% for
Qwen3-VL-8B-Instruct; on P4, the corresponding scores are
45.79\% and 25.58\%.
Its P2 accuracy also approaches the 71.48\% achieved by
Seed-2.0-Pro and Qwen3.7-Plus, narrowing the gap with the
leading closed-source systems in evidence selection.These results suggest that the Thinking variant's reasoning capabilities contribute to its higher accuracy in evidence selection and trigger recognition.


\subsection{Joint Analysis}
\begin{figure}[t]
    \centering
    \includegraphics[width=\textwidth]{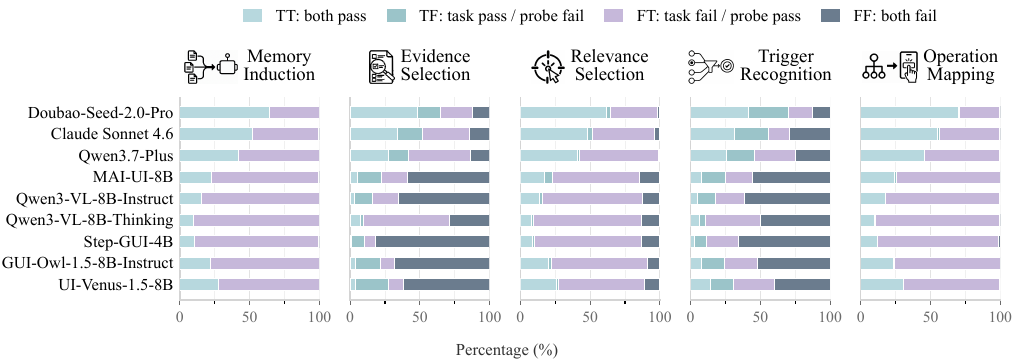}
    \caption{
\textbf{Joint outcomes between end-to-end task execution and stage-wise diagnostic
probes.} For each system and diagnostic stage, we partition paired task-probe
outcomes into four categories: TT (both pass), TF (the task passes / the probe fails),
FT (the task fails / the probe passes), and FF (both fail). 
The distributions reveal how failures identified by individual diagnostic stages relate to failures
in complete task execution.
}
    \label{fig:joint-outcomes}
    \vspace{-2pt}
\end{figure}

For each of the nine directly evaluated agents, we pair every
task-level diagnostic result with the corresponding end-to-end outcome
under behavioral memory logs.
Since the diagnostic set is deduplicated across agents, a task included
due to one agent's failure may still be completed by another agent.
Figure~\ref{fig:joint-outcomes} partitions the eligible paired instances
into four joint outcomes: both task and probe pass (TT), the task passes but the probe fails (TF), the task fails but the probe passes (FT), and both task and probe fail (FF).

\paragraph{P2 and P4 failures frequently co-occur with task failures.}
P2 Evidence Selection and P4 Trigger Recognition show markedly
higher joint-failure proportions than the other diagnostic stages.
Qwen3-VL-8B-Instruct fails both the task and its associated probe
on 65.5\% of P2 pairs and 61.7\% of P4 pairs.
The corresponding proportions are 68.1\% and 51.9\% for
GUI-Owl-1.5-8B-Instruct, and 82.1\% and 66.0\% for Step-GUI-4B.
These results connect the weaknesses identified by the independent
probes to failures on the corresponding GUI tasks, reinforcing
evidence selection and trigger recognition as important targets
for improvement.


\subsection{Controlled Evidence Intervention}

To test whether P2 Evidence Selection is an actionable bottleneck,
we evaluate Seed-2.0-Pro, Qwen3.7-Plus, Qwen3-VL-8B-Thinking,
and Step-GUI-4B on a fixed subset of 85 tasks selected from the
265 preference-conditioned tasks.
Table~\ref{tab:c0-c2-results} compares three conditions:
\textbf{C0 Raw Logs} uses the original logs;
\textbf{C1 Oracle Evidence Labels} retains all entries and labels
each as \textit{supporting} or \textit{noise} for the target preference,
without changing its text or order;
\textbf{C2 Supporting Logs Only} removes all noise entries and
retains only supporting logs.
C1 tests whether explicitly identifying supporting and noise logs
helps execution, while C2 tests whether removing noise is more
effective than retaining and labeling it.
The task instruction, initial environment state, and success checker
remain fixed.

\paragraph{Evidence labeling can outperform noise removal.}
From C0 to C1, Seed-2.0-Pro improves from 76.47\% to 83.53\%,
a gain of 7.06 percentage points, and Qwen3.7-Plus improves
from 50.59\% to 56.47\%.
These gains show that explicitly distinguishing supporting logs
from noise can help agents complete personalized tasks.
However, replacing labeled logs with supporting-only logs (C2)
reduces their TSR to 81.18\% and 51.76\%, respectively.
For these systems, identifying which logs support the preference
is therefore more effective than simply discarding the rest.
A possible explanation is that labeled noise provides contrastive
cues about which historical behaviors should not guide the current
decision, whereas deleting these entries removes those cues.

\paragraph{Limited GUI capability may constrain intervention benefits.}
Qwen3-VL-8B-Thinking achieves 69.53\% on P2 and 45.79\%
on P4, but only 30.19\% on \textit{Basic} tasks and 12.07\%
on the full preference-conditioned set under behavioral logs
(Table~\ref{tab:e2e-results}).
On the 85-task intervention subset, its TSR remains low:
17.65\% under C0, 16.47\% under C1, and 18.82\% under C2.
This contrast suggests that limitations in GUI execution constrain
the gains from evidence guidance, even when evidence selection is relatively strong.
Step-GUI-4B also reaches only 18.82\% under both C1 and C2.
These results suggest that strong GUI execution is a necessary foundation
for personalized task success: when an agent's basic GUI capability is
limited, even accurate preference understanding may not translate into
successful end-to-end execution.

\begin{table}[t]
\centering

\caption{
Additional analyses on PrefDiag-Bench (TSR, \%).
(a) Performance across C0--C2 settings.
C0 uses the original logs;
C1 retains all logs and labels each as supporting or noise
for the target preference;
C2 retains only supporting logs.
(b) End-to-end TSR under behavioral memory logs using different
memory modules.
Best and second-best results in each column are highlighted
with darker and lighter purple shading, respectively.
}

\begin{subtable}[t]{0.47\textwidth}
\centering
\caption{Performance across C0--C2 settings.}
\label{tab:c0-c2-results}
\vspace{-2pt}
\small
\setlength{\tabcolsep}{4.0pt}
\renewcommand{\arraystretch}{1.10}

\resizebox{0.835\linewidth}{!}{%
\begin{tabular}{l|ccc|c}
\hline

\textbf{System}
&
\textbf{C0}
&
\textbf{C1}
&
\textbf{C2}
&
\textbf{AVG}
\\

\hline

Step-GUI-4B
&
14.12
&
18.82
&
18.82
&
17.25
\\

Qwen3-VL-Thinking-8B
&
17.65
&
16.47
&
18.82
&
17.65
\\

Qwen3.7-Plus
&
\secondcell{50.59}
&
\secondcell{56.47}
&
\secondcell{51.76}
&
\secondcell{52.94}
\\

Seed-2.0-Pro
&
\bestcell{76.47}
&
\bestcell{83.53}
&
\bestcell{81.18}
&
\bestcell{80.39}
\\

\hline
\end{tabular}%
}

\end{subtable}
\hfill
\begin{subtable}[t]{0.51\textwidth}
\centering
\caption{Results with different memory modules.}
\label{tab:memory-module-results}

\small
\setlength{\tabcolsep}{2.8pt}
\renewcommand{\arraystretch}{1.10}

\resizebox{\linewidth}{!}{%
\begin{tabular}{l|ccccc|c}
\hline

\textbf{Method}
&
\textbf{Outcome}
&
\textbf{Constraint}
&
\textbf{Ranking}
&
\textbf{Process}
&
\textbf{Interaction}
&
\textbf{AVG}
\\

\hline

Mem0
&
71.70
&
56.60
&
43.40
&
\secondcell{50.94}
&
54.72
&
55.47
\\

LangMem
&
60.38
&
\secondcell{58.49}
&
41.51
&
43.40
&
49.06
&
50.57
\\

Hindsight
&
\secondcell{79.25}
&
54.72
&
\bestcell{45.28}
&
37.74
&
\secondcell{60.38}
&
\secondcell{55.47}
\\

Ours
&
\bestcell{83.02}
&
\bestcell{62.26}
&
\secondcell{45.28}
&
\bestcell{60.38}
&
\bestcell{77.36}
&
\bestcell{65.66}
\\

\hline
\end{tabular}%
}

\end{subtable}

\end{table}
\subsection{Robustness to Memory Construction Methods}
To test whether the preference-type differences in
Section~\ref{sec:end-to-end-results} depend on our original
memory construction, we evaluate Seed-2.0-Pro on all 265
preference-conditioned tasks using alternative memory inputs
produced by Mem0~\citep{ChhikaraKASY25}, LangMem~\citep{langchain2025langmem}, and Hindsight~\citep{latimer-etal-2026-hindsight} from the same source
behavioral records.
The task instruction, initial environment state, success checker,
and execution settings remain fixed.
Table~\ref{tab:memory-module-results} reports TSR by preference type,
with the original behavioral memory logs included as \textsc{Ours}.
\paragraph{Preference-type differences persist across memory construction methods.}
Outcome remains the highest-scoring type, while Ranking and Process
are the two lowest-scoring types under all three methods.
With Mem0, TSR is 71.70\% on Outcome, compared with 43.40\%
on Ranking and 50.94\% on Process.
Hindsight shows the same broad contrast, reaching 79.25\% on
Outcome, 45.28\% on Ranking, and 37.74\% on Process.
Although the absolute scores and the ordering of Ranking and Process
change, their relative difficulty persists across the tested methods.
This consistency further supports our preference-conditioned task
design, as the observed differences across preference types persist
under alternative memory construction methods rather than being
specific to our original behavioral memory logs.

\section{Conclusion}
\label{sec:conclusion}

We introduce PrefDiag-Bench to provide a more systematic and
comprehensive view of personalization in mobile GUI agents.
By categorizing user preferences into five types and combining
this categorization with a stage-wise diagnostic framework,
it examines how preferences affect agent execution, when they
should take effect, and where capability gaps arise.
Our evaluation shows that overall task success alone can obscure
distinct strengths and weaknesses across preference types and
stages of preference application.
Connecting these capability-level assessments with end-to-end
outcomes provides a basis for more informative evaluation and
targeted interventions.
We hope PrefDiag-Bench supports the development of GUI agents
that not only understand user preferences but also apply them
reliably throughout task execution.

\section*{AI Use Statement}

In this work, we used generative AI tools (GPT-5.6 Sol, Extra high) to assist with synthetic data
generation and translation, as well as language editing and polishing.
AI-generated synthetic data were manually reviewed and revised by the
authors before being included in the benchmark. All other AI-assisted
content was also reviewed by the authors. The authors take full
responsibility for the final content of this work.

\bibliography{references}
\bibliographystyle{assets/iclr2027_conference}

\appendix
\clearpage
\section{Dataset Details and Examples}
\label{sec:appendix-details}

This appendix presents the task catalog, paired user contexts, and
examples of P1--P5. Task families are identified as \textsc{PrefD}~01--53.

\subsection{Task Catalog and Examples}
\label{sec:appendix-task-catalog}

\paragraph{Task sources and catalog.}
Table~\ref{tab:appendix-task-catalog} lists the 53 \textit{Basic} tasks
adapted from KnowU-Bench~\citep{chen2026knowubenchinteractiveproactivepersonalized},
with one row per family.
We inspect each task's instruction, initial state, and completion
conditions, adapting candidates, delivery information, messages,
calendar events, and local information pages where needed.
Each family has five preference-conditioned variants, yielding 318
end-to-end instances. Profiles and behavioral logs provide alternative
contexts for the same 265 personalized tasks.
The catalog lists the applications used by each \textit{Basic} task;
variants require additional applications, such as for cross-platform
comparison.

\begingroup
\footnotesize
\setlength{\tabcolsep}{4pt}
\renewcommand{\arraystretch}{1.17}
\setlength{\LTleft}{0pt plus 1fill}
\setlength{\LTright}{0pt plus 1fill}
\begin{longtable}{@{}>{\raggedright\arraybackslash}p{0.11\textwidth}
>{\raggedright\arraybackslash}p{0.16\textwidth}
>{\raggedright\arraybackslash}p{0.16\textwidth}
>{\raggedright\arraybackslash}p{\dimexpr0.57\textwidth-24pt\relax}@{}}
\caption{Catalog of the 53 Basic tasks. Each family also contains five
preference-conditioned variants. The application column lists the Basic
task's applications, not every application available in its environment.}
\label{tab:appendix-task-catalog}\\
\toprule
Family ID & Application(s) & Domain & Base task description \\
\midrule
\endfirsthead
\multicolumn{4}{l}{\tablename~\thetable\ (continued)}\\
\toprule
Family ID & Application(s) & Domain & Base task description \\
\midrule
\endhead
\midrule
\multicolumn{4}{r}{Continued on the next page}\\
\endfoot
\bottomrule
\endlastfoot
PrefD~01 & Taodian & Commerce & Purchase one case of canned sugar-free Coca-Cola, enter the specified recipient details, and submit the order. \\
PrefD~02 & Taodian & Commerce & Purchase white everyday commuting sneakers in size 42, enter the specified recipient details, and submit the order. \\
PrefD~03 & Taodian & Commerce & Purchase an Apple MacBook Air with an M-series processor, at least 16\,GB RAM, and a price no greater than CNY~10,000. \\
PrefD~04 & Taodian & Commerce & Remove all clothing with a unit price above CNY~140 from the cart; exclude shoes and preserve all other items. \\
PrefD~05 & Taodian & Commerce & Remove black clothing and all bodysuits from the cart while leaving other products unchanged. \\
PrefD~06 & Chilemei & Commerce & Order a light Japanese- or Beijing-style lunch without peanuts, costing at most CNY~35, and complete simulated payment. \\
PrefD~07 & Chilemei & Commerce & Order kung pao chicken with rice, add the note ``no peanuts, less chili, extra rice,'' and complete simulated payment. \\
PrefD~08 & Taodian & Commerce & Purchase a sofa and add an order note requesting shipment by SF Express. \\
PrefD~09 & Chrome & Information retrieval & Visit the AlphaXiv and Hugging Face Papers trending-paper pages and finish on Hugging Face Papers. \\
PrefD~10 & Chrome & Information retrieval & Search for today's Hangzhou weather, inspect precipitation probability and temperature extremes, and remain on the result page. \\
PrefD~11 & Maps & Navigation & Plan a sunny-day bicycle commute from Zhejiang University's Zijingang campus to its science park; report the route, distance, and duration. \\
PrefD~12 & Maps & Navigation & Plan the campus-to-science-park commute in rainy weather, minimizing exposure to rain, and report a travel plan. \\
PrefD~13 & Maps & Navigation & Plan a taxi trip from the science park to Zijingang campus at 23:30, after bus and metro services have ended. \\
PrefD~14 & Maps; Mattermost & Navigation & Plan the fastest campus-to-science-park route and post a message in Town Square reporting an approximately 20-minute delay. \\
PrefD~15 & Calendar; Mattermost & Office & Create a project-discussion meeting for 14:00--15:00 on February 10, 2026, and notify the team-chat channel. \\
PrefD~16 & Messages; Calendar & Office & Accept Alex's 10:30--11:00 project-sync invitation on February 10, 2026, reply by SMS, and add the meeting to the calendar. \\
PrefD~17 & Files; Docreader & Office & Find and open the specified Agent Learning via Early Experience PDF in Documents before the Product Review meeting. \\
PrefD~18 & Clock & Office & Create an 08:30 Saturday-and-Sunday alarm with Bright Morning; retain the default ringtone if that choice is unavailable. \\
PrefD~19 & Clock & Office & Turn off the existing 07:30 alarm before the weekend. \\
PrefD~20 & Settings & System settings & Enable Battery Saver at 15\% battery and remain on the battery-settings page. \\
PrefD~21 & Settings & System settings & Mute media volume after Bluetooth headphones disconnect, without changing Bluetooth settings. \\
PrefD~22 & Settings & System settings & Enable Do Not Disturb or silence the phone for a 09:00--11:30 focused-work session. \\
PrefD~23 & Messages; Contacts & Communication & Save the sender of the new-number message as Bob, using phone number 5550199. \\
PrefD~24 & Messages & Communication & Send Mom a birthday SMS containing ``Happy Birthday'' or its Chinese equivalent. \\
PrefD~25 & Messages & Communication & Send Mom the specified message requesting CNY~2,000 for living expenses. \\
PrefD~26 & Messages; Contacts & Communication & Save the urgent Project Office sender as a contact and mark the contact as a favorite. \\
PrefD~27 & Messages & Communication & Block the specified phishing-message sender without replying or opening the message's link. \\
PrefD~28 & Mattermost & Communication & Acknowledge the Cluster-A server-error alert in town-square and state that it is being investigated. \\
PrefD~29 & Mattermost & Communication & Post the specified one-day sick-leave notice in Town Square. \\
PrefD~30 & Mail; Files & Communication & Email the dean with subject ``Weekly Report,'' the specified body, and the Weekly\_Report.pdf attachment. \\
PrefD~31 & Files & Office & Sort the Download folder by modification time, newest first. \\
PrefD~32 & Gallery; Mail & Office & Select the flower photos and email them to Kevin with the specified message. \\
PrefD~33 & Gallery; Mastodon & Social media & Publish the HuggingGPT architecture image from the gallery with a short caption using the logged-in account. \\
PrefD~34 & Mastodon; Mattermost & Social media & Find an AI- or machine-learning-related post and share its content and original link in Town Square. \\
PrefD~35 & Mastodon & Social media & Publish the specified weekend-hiking update with public visibility. \\
PrefD~36 & Gallery & Office & Delete screenshots older than seven days while preserving those from the most recent seven days. \\
PrefD~37 & Taodian & Commerce & Remove electronic and electronics-related products from the cart while preserving other categories. \\
PrefD~38 & Chrome; Mail & Information retrieval & Use the local search page to find recent AI news, then email Kevin an AI-titled message with the required opening and a summary. \\
PrefD~39 & Contacts; Phone & Communication & Find the contact Son (Qiang) and initiate a phone call. \\
PrefD~40 & Mattermost & Communication & Post the specified end-of-workday farewell in town-square. \\
PrefD~41 & Settings & System settings & Switch the phone to the system dark/night theme while browsing Mastodon late at night. \\
PrefD~42 & Calendar; Files; Docreader & Office & Find the document specified in the Product Review calendar event and open it for pre-meeting reading. \\
PrefD~43 & Taodian & Commerce & Purchase one bag of Mingqian Yirgacheffe medium-dark-roast whole coffee beans and submit the order with the specified delivery details. \\
PrefD~44 & Chilemei & Commerce & Order jasmine green milk tea with 30\% sugar, no ice, and oat milk; request a sealed cup and a separate straw, then complete simulated payment. \\
PrefD~45 & Maps & Navigation & Plan a trip from Aokesi, Hangzhou, to Hangzhou East railway station with two large suitcases and about 65 minutes before ticket inspection. \\
PrefD~46 & Messages; Calendar & Office & Politely decline Lin Na's Saturday 19:00--21:00 dinner invitation because it conflicts with family time; do not add the dinner to the calendar. \\
PrefD~47 & Messages; Contacts & Communication & Save the parent-meeting notice sender as Duoduo's class teacher and mark the contact as a favorite. \\
PrefD~48 & Gallery; Mastodon & Social media & Publish the specified photo of the user's pet Juzi sunbathing by a window, with the prescribed caption. \\
PrefD~49 & Files & Office & Move report.pdf, plan.docx, and budget.xlsx from Download to the work-documents folder; leave photo.jpg, movie.mp4, and app.apk in place. \\
PrefD~50 & Clock & Office & Create an enabled 07:00 Saturday-and-Sunday alarm using Cesium, the system-default alarm sound. \\
PrefD~51 & Chrome & Information retrieval & Look up today's Beijing weather and maximum temperature in the morning, and report them to the user. \\
PrefD~52 & Contacts; Phone & Communication & Find Grandma in Contacts and initiate an after-dinner phone call. \\
PrefD~53 & Settings & System settings & Open Settings and enable Battery Saver when the phone's battery is running low. \\
\end{longtable}
\endgroup

\paragraph{Task-family example.}
Table~\ref{tab:appendix-family-example} presents the six instances of
\textsc{PrefD}~02.
The \textit{Basic} instruction specifies the shoes and delivery details,
whereas the Outcome variant obtains the user's usual choice and
recipient details from context.
The remaining variants retain an explicit purchase goal while introducing
a budget, store priority, procedure, or confirmation requirement.
Each variant is a separate task instance.

\begingroup
\small
\setlength{\tabcolsep}{4pt}
\renewcommand{\arraystretch}{1.12}
\begin{longtable}{@{}>{\raggedright\arraybackslash}p{0.12\textwidth}
>{\raggedright\arraybackslash}p{\dimexpr0.44\textwidth-8pt\relax}
>{\raggedright\arraybackslash}p{\dimexpr0.44\textwidth-8pt\relax}@{}}
\caption{User inputs for the six instances in \textsc{PrefD}~02, shown as
complete English translations with the application names normalized as in
Section~\ref{sec:appendix-task-catalog}. In the behavioral-log
condition, the preference is replaced by the corresponding logs.}
\label{tab:appendix-family-example}\\
\toprule
Variant & User Instruction & Explicit User Preference \\
\midrule
\endfirsthead
\multicolumn{3}{l}{\tablename~\thetable\ (continued)}\\
\toprule
Variant & User Instruction & Explicit User Preference \\
\midrule
\endhead
\midrule
\multicolumn{3}{r@{}}{Continued on the next page}\\
\endfoot
\bottomrule
\endlastfoot
Basic &
Open Taodian, search for a pair of white everyday commuting sneakers,
choose size 42, enter Zhang Wei as the recipient, 13800138000 as the phone
number, and Aokesi, Dongcheng District, Beijing, as the address, then place
the order. &
None. \\
\addlinespace[4pt]
Outcome &
Help me buy a pair of the sneakers I usually wear for commuting to work
on Taodian; just place the order. &
The user likes simple white sneakers for commuting to and from work,
dislikes flashy colors and high-top styles, and wears size 42. The user's
name is Zhang Wei, phone number is 13800138000, and default delivery address
is Aokesi, Dongcheng District, Beijing. \\
\addlinespace[4pt]
Constraint &
Help me search Taodian for a pair of white everyday commuting sneakers,
choose size 42, enter Zhang Wei as the recipient, 13800138000 as the phone
number, and Aokesi, Dongcheng District, Beijing, as the address, then place
the order. &
The user values cost-effectiveness when buying commuting shoes and does
not like buying expensive ones; the price of a pair must not exceed
CNY~300. If only options above CNY~300 are available, the user must first
be told why; an order must not be placed directly. \\
\addlinespace[4pt]
Ranking &
Help me search Taodian for a pair of white everyday commuting sneakers,
choose size 42, enter Zhang Wei as the recipient, 13800138000 as the phone
number, and Aokesi, Dongcheng District, Beijing, as the address, then place
the order. &
The user cares strongly about authenticity when buying shoes and prefers
brand official flagship stores (official direct sales). If the same shoe
is available from both an official flagship store and an ordinary store,
the user usually prioritizes the official flagship store and considers
an ordinary store only when the official flagship store is out of stock. \\
\addlinespace[4pt]
Process &
Help me search for a pair of white everyday commuting sneakers, choose
size 42, enter Zhang Wei as the recipient, 13800138000 as the phone number,
and Aokesi, Dongcheng District, Beijing, as the address, then place the
order. &
The user is conscious about price when buying shoes and dislikes placing an
impulsive order on a single platform. The user habitually compares the
same item's prices on Taodian and JD before ordering from the cheaper
platform. \\
\addlinespace[4pt]
Interaction &
Help me search Taodian for a pair of white everyday commuting sneakers,
choose size 42, enter Zhang Wei as the recipient, 13800138000 as the phone
number, and Aokesi, Dongcheng District, Beijing, as the address, then place
the order. &
The user is fairly cautious and dislikes automated actions making decisions
on their behalf. For any action that places an order, the user wants to be
asked for confirmation beforehand and does not want an order placed
directly without confirmation. \\
\end{longtable}
\endgroup

\subsection{User-Context Construction and Annotation}
\label{sec:appendix-user-context}

\begin{figure}[t]
    \centering
    \includegraphics[width=\linewidth]{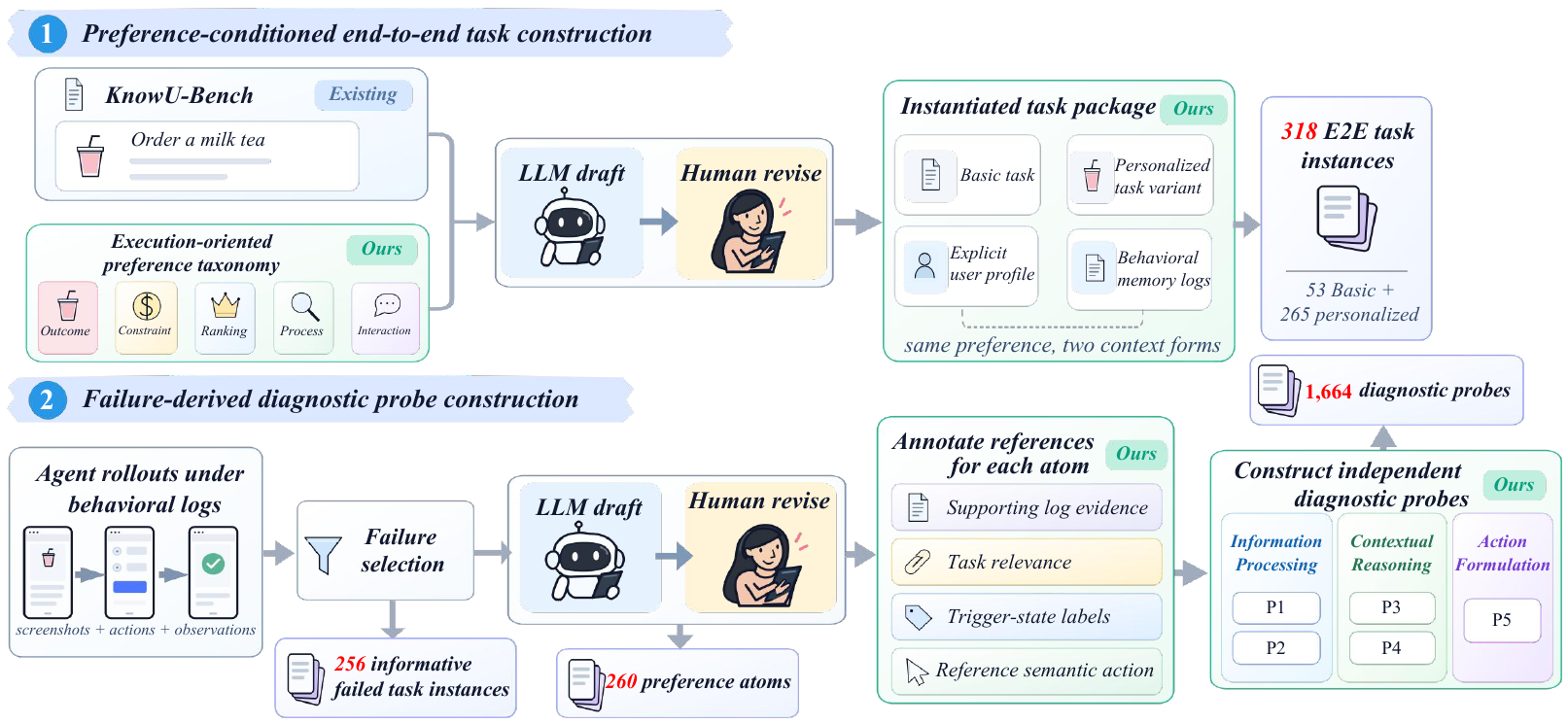}
    \caption{
    \textbf{Construction pipeline of PrefDiag-Bench.}
    \textit{Top:} Task construction with paired profiles and behavioral logs.
    \textit{Bottom:} Failure selection, preference-atom extraction,
    and diagnostic-probe construction.
    Both stages combine model-assisted drafting with human revision.
    }
    \label{fig:construction-pipeline}
\end{figure}

Figure~\ref{fig:construction-pipeline} summarizes the overall
construction process. We detail below how explicit user profiles,
behavioral memory logs, and their reference annotations are produced.

\paragraph{Profile drafting and review.}
For each personalized task, we specify the intended preference
and its applicable conditions. An LLM drafts an explicit user profile,
which GPT-5.6 Sol checks and revises before human reviewers examine
every profile. Reviewers check the requested behavior, applicable
conditions, and task-specific details against the intended preference.
The reviewed profile then serves as the basis for log synthesis.

\paragraph{Behavioral-log synthesis.}
Using the reviewed profile, an LLM generates synthetic records of past
choices, corrections, confirmations, and related activities.
Supporting records provide evidence for the target preference,
while distractors may share an application, product category, or
recipient without establishing the same requirement.
Contextual exceptions are retained where present.
The logs undergo GPT-5.6 Sol revision and item-by-item human review
alongside the profile.
Reviewers check that both forms convey compatible requirements
and that distractors remain plausible historical records rather
than explicitly declaring their irrelevance.
The instruction, initial environment state, and success checker
remain fixed across the paired context forms.

\paragraph{Reference annotation and validation.}
We link each preference atom to its supporting log entries and
annotate task applicability, trigger conditions, and the expected
behavior for diagnostic construction.
Reviewers check evidence support against the actual record content.
For the confirmation example in
Section~\ref{sec:appendix-context-example}, a recorded
ask--confirm--submit sequence supports the preference, whereas
store selection or price comparison does not.
When inconsistencies are found, the source records are revised
and the affected context pairs and probes are rechecked.
Automated checks validate identifiers, required fields, and evidence
references.

\Needspace{8\baselineskip}
\subsubsection{Paired-Context Example: \textsc{PrefD}~02 / Interaction}
\label{sec:appendix-context-example}

\textbf{Reference preference.} Request confirmation before submitting
 the shoe order.

\textbf{Task instruction.}
\begin{quote}
Help me search Taodian for a pair of white everyday commuting sneakers,
choose size 42, enter Zhang Wei as the recipient, 13800138000 as the phone
number, and Aokesi, Dongcheng District, Beijing, as the address, then place
the order.
\end{quote}

\textbf{Explicit user profile.}
\begin{quote}
The user is fairly cautious and dislikes automated actions making decisions
on their behalf. For any action that places an order, the user wants to be
asked for confirmation beforehand and does not want an order placed
directly without confirmation.
\end{quote}

\textbf{Behavioral memory logs.}
\label{ex:appendix-log-input}
\begin{quote}
\small
\textbf{L1.} During a previous automated shoe purchase, the agent asked,
``Do you confirm ordering this white pair in size 42?'' before clicking
the order button. The order was submitted only after the user confirmed.

\textbf{L2.} The user once canceled an automatically placed clothing order
and added that future orders should be preceded by a question, rather than
submitted directly.

\textbf{L3.} The user bought three pairs of white ankle socks in Taodian
for CNY~19.9, with delivery to Aokesi, Dongcheng District, Beijing.

\textbf{L4.} For a shoe-care purchase, the user selected a cleaning product
from an ordinary store because the official store did not have the same
item in stock.

\textbf{L5.} When buying a backpack, the user checked the same model's
prices in Taodian and Jingdian, then placed the order in Jingdian.
\end{quote}

\textbf{Reference evidence.} L1 and L2 support confirmation before order
submission: L1 records an ask--confirm--submit sequence, and L2 records a
correction after automatic ordering. L3 concerns a socks purchase, L4 store
selection under a stock exception, and L5 cross-platform price comparison.
These records do not directly support the confirmation requirement.
The record identifiers L1--L5 are presentation labels; the reference
evidence is not part of the ordinary behavioral-log input.

\subsection{Diagnostic Probe Examples}
\label{sec:appendix-probe-examples}

The diagnostic set contains 1,664 items: 260 P1, 260 P2, 325 P3,
601 P4, and 218 P5.
The examples below correspond to \textsc{PrefD}~02 / Interaction.
Each probe is queried independently with the specified inputs, without
passing predictions between probes.
Shared inputs are printed once and cross-referenced, but supplied in
full in each request. Reference answers and explanations are presented
separately from model inputs.

\Needspace{8\baselineskip}
\subsubsection{P1: Log-to-Memory Induction}
\label{sec:appendix-p1-example}

\textbf{Example ID.} P1--PrefD~02--Interaction.

\textbf{Input.} The five logs L1--L5 in
Section~\ref{sec:appendix-context-example}, supplied in full, together
with the Question.

\textbf{Question.} Based only on the memory logs above, which statement
best summarizes the user's stable communication and confirmation
preferences when this kind of task is automated?

\begin{description}
\item[A.] Ask whether the user is satisfied after submitting a clothing or
shoe order, and try to remedy it if the user disagrees.
\item[B.] Confirm the style only at the start of the search; the automation
may decide whether to place the final order on its own.
\item[C.] Ask every time the product or size changes; filtering cannot
proceed continuously.
\item[D.] Search and filtering may proceed continuously, but the user
should have a final decision opportunity before a clothing or shoe order
is placed.
\end{description}
\textbf{Response format.} One option ID, A--D.

\textbf{Reference answer: D.}
L1 and L2 establish confirmation before order submission, rather than
confirmation after ordering, only at search initiation, or at every
product/size change.

\Needspace{8\baselineskip}
\subsubsection{P2: Evidence Selection}
\label{sec:appendix-p2-example}

\textbf{Example ID.} P2--PrefD~02--Interaction.

\textbf{Input.} The known preference is: ``Search and filtering may proceed
continuously, but the user should have a final confirmation opportunity
before a shoe order is submitted.''

\textbf{Question.} The scope above only limits where the preference
applies and does not itself need to be established from the logs. Which
memory logs directly support the preference being verified? Select all
that apply.

\begin{description}
\item[A.] The user once canceled an automatically placed clothing order
and added that future orders should be preceded by a question, rather
than submitted directly.
\item[B.] Several previous personal shopping orders used Zhang Wei,
13800138000, and Aokesi, Dongcheng District, Beijing, as the default
delivery details.
\item[C.] The user bought three pairs of white ankle socks in Taodian
for CNY~19.9, with delivery to Aokesi, Dongcheng District, Beijing.
\item[D.] During a previous automated shoe purchase, the agent asked,
``Do you confirm ordering this white pair in size 42?'' before clicking
the order button. The order was submitted only after the user confirmed.
\item[E.] For a shoe-care purchase, the user selected a cleaning product
from an ordinary store because the official store did not have the same
item in stock.
\item[F.] When buying a backpack, the user checked the same model's prices
in Taodian and Jingdian, then placed the order in Jingdian.
\item[G.] During a shoe purchase, the user first opened a colorful
high-top style, then went back and selected white low-top commuting shoes,
adding the note ``for work; nothing too flashy.''
\item[H.] In the user's most recent personal commuting-shoe purchase,
the user chose simple white low-top sneakers in size 42, with Zhang Wei,
13800138000, and Aokesi, Dongcheng District, Beijing, as the delivery details.
\end{description}
\textbf{Response format.} A JSON array of selected option IDs in alphabetical order.

\textbf{Reference answer: \texttt{["A", "D"]}.}
A states the confirmation rule through a correction, and D records the
corresponding behavior. The other options concern delivery, product choice,
store choice, or price comparison. The selected set must match the reference
exactly.

\subsubsection{P3: Relevance Selection}
\label{sec:appendix-p3-example}

\textbf{Example ID.} P3--PrefD~02--Interaction.

\textbf{Input.} The complete task instruction from
Section~\ref{sec:appendix-context-example} and the preference candidates
below.

\textbf{Question.} Which known user preference should affect planning or
execution at an appropriate point in the current task? Consider the whole
task.

\begin{description}
\item[A.] None of these preferences should affect the current task.
\item[B.] When ordering a personal weekday lunch for delivery, finish
choosing food and entering the address first, then summarize the restaurant,
dishes, and amount and obtain one confirmation before payment.
\item[C.] When purchasing a laptop for the user's work, present the model,
price, delivery location, and other key risk information together before
ordering, and submit only after confirmation.
\item[D.] When purchasing everyday commuting shoes for the user, search
and filtering may proceed continuously, but the user should have a final
confirmation opportunity before a shoe order is submitted.
\end{description}
\textbf{Response format.} One option ID, A--D.

\textbf{Reference answer: D.}
D matches the current shoe-purchase task. B and C specify confirmation
requirements for lunch and laptop purchases. The question concerns relevance
to the task, not the current trigger state.

\clearpage
\subsubsection{P4: Trigger Recognition}
\label{sec:appendix-p4-example}

\textbf{Example ID.} P4--PrefD~02--Interaction--Active.

\begin{wrapfigure}{r}{0.36\linewidth}
  \vspace{-0.7\baselineskip}
  \centering
  \includegraphics[
    width=0.95\linewidth,
    height=0.34\textheight,
    keepaspectratio
  ]{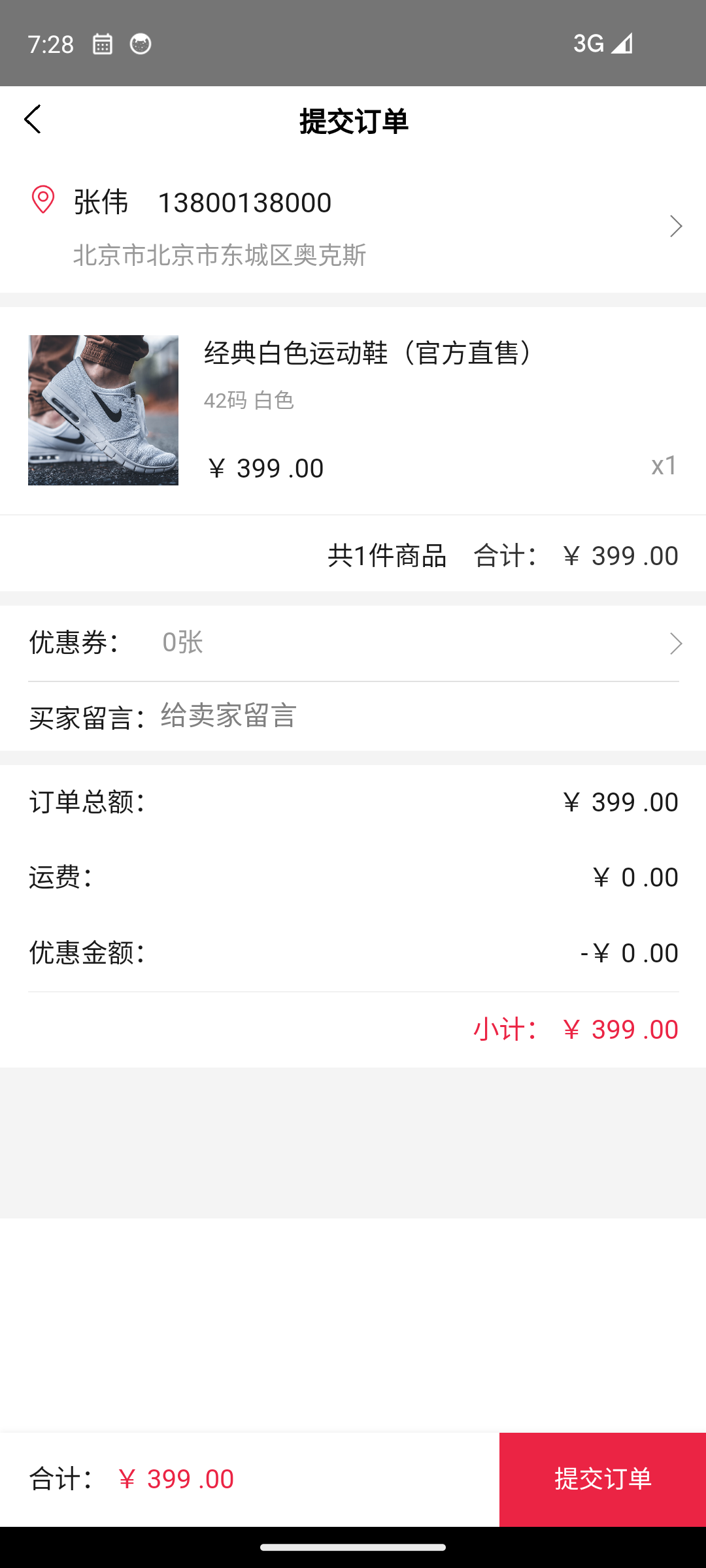}
  \caption{GUI observation used in the P4 and P5 examples.
  The screen shows the pending shoe order and the red
  ``Submit order'' button. Interface text is translated for
  the reader and is not additional model input.}
  \label{fig:appendix-confirmation-state}
  \vspace{-0.4\baselineskip}
\end{wrapfigure}

\textbf{Input.} The complete task instruction from
Section~\ref{sec:appendix-context-example}, the preference below, the prior
action history, and the screenshot in
Figure~\ref{fig:appendix-confirmation-state}.

\textbf{Given preference.} Search and filtering may proceed continuously,
but the user should have a final confirmation opportunity before a shoe
order is submitted.

\textbf{Prior action history.}

\noindent
1.\ Searched for and selected ``Classic White Sneakers (Official Direct).''\\
2.\ Selected size 42.\\
3.\ Entered Zhang Wei, 13800138000, and Aokesi, Dongcheng District,
Beijing.\\
4.\ Returned to the order-submission page.

\medskip

\textbf{Question.} Given the current screenshot and the preceding objective
action history, which stage is this preference at in the current task state?

\begin{description}
\item[A.] The relevant decision has already been completed, or the trigger
window for this preference has passed.

\item[B.] The decision point at which this preference needs to be applied
has not yet been reached.

\item[C.] The current state is at the trigger point for this preference.
\end{description}

\textbf{Response format.} One option ID, A--C.

\textbf{Reference answer: C (\texttt{active\_now}).}
In the reference item, order submission is pending after the product and
delivery details have been specified, placing the interaction at the
final-confirmation point. The screenshot precedes action 31 of the reference
trajectory. Only preceding actions are supplied: future actions, model
reasoning, and the reference label are excluded.
The other classes are \texttt{not\_active\_yet} and
\texttt{no\_longer\_active}.

\Needspace{8\baselineskip}
\subsubsection{P5: Operation Mapping}
\label{sec:appendix-p5-example}

\textbf{Example ID.} P5--PrefD~02--Interaction.

\textbf{Input.} The task instruction, preference, prior action history, and
screenshot used in the P4 example are supplied again in a separate request.
The additional condition states that the preference is active and the
relevant decision remains pending; it is not obtained from the model's
P4 prediction.

\textbf{Question.} Which next-step strategy best follows this preference
at the current moment?

\begin{description}
\item[A.] Summarize the current shoe, size 42, and pending order; ask the
user to confirm submission and wait for an explicit reply before deciding
whether to click ``Submit order.''
\item[B.] Click ``Submit order'' directly. The original instruction already
requested an order, so no further confirmation is needed.
\item[C.] Click ``Submit order'' first, then tell the user the result on
the payment page and ask whether it is acceptable.
\item[D.] Tell the user ``I am about to place the order,'' then click
``Submit order'' immediately without waiting for a reply.
\end{description}
\textbf{Response format.} One option ID, A--D.

\textbf{Reference answer: A.}
A requests confirmation before submission and waits for the user's reply.
B omits confirmation, C asks after submission, and D replaces confirmation
with a notice. The item tests a semantic strategy, not screen coordinates
or an action-token sequence.


\section{Experimental Details}
\label{sec:appendix-experiments}

\subsection{Environment and Agent Configuration}
\label{sec:appendix-environment}

\paragraph{Environment and task inputs.}
We reuse the Android environment and evaluation infrastructure of
KnowU-Bench,
with task-specific initial data and checkers for PrefDiag-Bench.
Each episode starts from its prescribed application state, with agent
history reset between tasks. Episodes allow at most 65 environment
steps with a one-second inter-step wait, and can end earlier when the
agent reports completion or inability to proceed.
Open-source models are served on eight NVIDIA RTX 4090 GPUs.
The 53 \textit{Basic} tasks receive no additional user context.
Each of the 265 personalized tasks receives either its explicit user
profile or its behavioral memory logs. 
The instruction and context remain available throughout execution,
alongside the current observation, adapter-specific history, and any
platform reply. Adapters retain native observation and action formats
while mapping actions and coordinates to the common environment interface.

\paragraph{Decoding and workflow interfaces.}
The Qwen, MAI-UI, Step-GUI, GUI-Owl, and UI-Venus adapters use
temperature 0. Seed-2.0-Pro uses temperature 0.7, top-$p=0.9$,
and a 4,096-token response limit; its adapter retains the three most
recent screenshots alongside textual history.
Claude's adapter does not explicitly set temperature.
M3A and T3A use GPT-5.4 with temperature 0, a 2,048 token action-response
limit, and up to 20 objective action summaries.
Both workflows use the same task contexts, step budget, and checkers
as the directly evaluated agents.

\subsection{Scoring and Task--Probe Association}
\label{sec:appendix-scoring}

\paragraph{End-to-end success.}
Let $r_i\in[0,1]$ be the native checker score for task instance $i$.
Binary success and TSR are defined as
\[
s_i=\mathbf{1}[r_i=1],
\qquad
\mathrm{TSR}=\frac{100}{|\mathcal{V}|}\sum_{i\in\mathcal{V}}s_i,
\]
where $\mathcal{V}$ contains tasks with valid checker results and TSR is
reported as a percentage. Checkers inspect application state, database or callback
records, and required action and interaction traces. Any accompanying
natural-language reason is not used as a separate success classifier.
The personalized AVG is the macro-average of the five preference-type
TSRs. 

\paragraph{Diagnostic accuracy.}
Each P1--P5 item is answered in an independent request, without passing
model predictions between probes. Single-choice items require the
reference option; P2 requires an exact match to the complete reference
option set.
For task--preference instance $i$ and probe type $k$, let
$\mathcal{Q}_{i,k}$ contain all associated canonical questions, including
those from multiple atoms or GUI states. The task-level result is
\[
a_{i,k}=\prod_{q\in\mathcal{Q}_{i,k}}
\mathbf{1}[\hat y_q=y_q],
\]
where $\hat y_q$ and $y_q$ are the predicted and reference answers.
Thus, answering two of three P4 items correctly gives a task-level
result of 0, not $2/3$.
Groups with any missing, truncated, or unparseable response are excluded,
and their coverage is recorded.

\paragraph{Paired outcomes.}
We join diagnostic and end-to-end results by model, task-family ID,
and preference type. Each eligible pair contributes once per probe type.
TT, TF, FT, and FF denote $(s_i,a_{i,k})=(1,1),(1,0),(0,1),(0,0)$,
respectively. Pooled proportions are calculated from valid pair counts,
not by averaging preference-wise percentages.

\subsection{Evidence Intervention}
\label{sec:appendix-intervention}

\paragraph{Task selection.}
The fixed subset contains 85 task--preference pairs from 43 families,
with 17 pairs per preference type: 15 P2-target pairs, one downstream
control, and one retention control. The totals are therefore 75, 5,
and 5 pairs, respectively.
Selection uses pre-intervention diagnostic and behavioral-log execution
records. A P2-target pair requires at least one model that fails both
P2 and the task, and at least one model that completes the task.
Downstream controls have correct P2 answers but failed execution;
retention controls have successful diagnostic and execution outcomes.
Every selected pair has an audited supporting-log mapping and at least
one distractor. This P2-enriched subset is shared across all four models
and all conditions, and selection does not use intervention outcomes.

\paragraph{Input conditions.}
C0 supplies the behavioral memory logs without evidence labels.
C1 preserves the same entries, text, and order, adding the prefix
\texttt{[Relevant evidence]} or \texttt{[Irrelevant context]}
according to the reference evidence annotation.
C2 retains only supporting entries in their original order, without
labels. 
All three conditions are executed anew with the same instruction,
initialization, and checker.

\subsection{Memory-Module Configurations}
\label{sec:appendix-memory-modules}

Mem0, LangMem, and Hindsight process the same frozen synthetic
historical episodes for each of the 265 task--preference pairs.
Each task is processed independently: we initialize
a fresh memory store for the pair, ingest only its associated episodes
in chronological order, and discard the store after evaluation.
No memory state is shared across tasks, families, or preference variants.
The reference profile, preference atom, evidence-role labels, and checker
are not supplied to the memory modules.
Table~\ref{tab:appendix-memory-settings} lists their operations and
configurations. The resulting memory text replaces Seed-2.0-Pro's
behavioral-log context without any reference-based correction of
omissions, merging errors, or over-generalizations.
Task initialization, interaction settings, the max step, and
checkers remain unchanged.
\begin{table}[!htbp]
\centering
\caption{Memory construction and retrieval settings.
All modules use GPT-5.4-mini for their LLM operations and an isolated
store for each task--preference pair. The returned memory text
is supplied to the downstream GUI agent.}
\label{tab:appendix-memory-settings}
\begingroup
\small
\setlength{\tabcolsep}{5pt}
\renewcommand{\arraystretch}{1.14}
\begin{tabularx}{\linewidth}{@{}>{\raggedright\arraybackslash}p{0.18\linewidth}
>{\raggedright\arraybackslash}X
>{\raggedright\arraybackslash}X@{}}
\toprule
\textbf{Module} & \textbf{Memory operation and agent input}
& \textbf{Configuration} \\
\midrule
Mem0\newline
\texttt{mem0ai} 2.0.17
& Sequential \texttt{add(infer=True)} extracts and consolidates memories.
Scoped \texttt{get\_all} returns the final records (limit 100);
all returned text is used, without task-conditioned search.
& Local Qdrant; \texttt{bge-small-zh-v1.5} embeddings (512 dimensions).
Temperature 0, top-$p=0.1$, and 2,000 output tokens. \\
\addlinespace[4pt]
LangMem\newline
0.0.30
& \texttt{create\_memory\_manager} updates a memory list episode by episode,
carrying the previous list forward. 
& Default memory instructions and \texttt{Memory(content)} schema;
inserts and updates enabled, deletes disabled.
Temperature 0, top-$p=0.1$, and 2,000 output tokens. \\
\addlinespace[4pt]
Hindsight\newline
0.9.0
& \texttt{retain} ingests episodes into a fresh bank.
\texttt{recall} uses the current GUI instruction as its query after
ingestion. Returned text is supplied in native order.
& \texttt{bge-m3} embeddings and \texttt{bge-reranker-v2-m3} reranking.
Recall budget \texttt{mid}, 4,096 tokens; world, experience, and observation
memories. \\
\bottomrule
\end{tabularx}
\endgroup
\end{table}

\section{Preference Application in GUI Tasks}
\label{sec:appendix-preference-cases}

The following examples illustrate how preferences affect GUI decisions.
\textit{Basic} uses the task instruction alone; \textit{Personalized}
adds behavioral memory logs. Instructions and preferences are summarized
below. Each figure aligns corresponding decision stages and retains
the original step numbers.

\subsection{Choosing an Official Store (\textsc{PrefD}~03, Ranking)}
\label{sec:appendix-case-laptop}

\noindent\textbf{Instruction.}
Buy an Apple MacBook Air M-series laptop in Taodian with at least
16\,GB of memory and a price of at most CNY~10,000. Submit the order
using the specified delivery details.

\noindent\textbf{Preference.}
Prefer an official store, even at a higher price. Use an ordinary
store only when the official option is out of stock.

\begin{figure}[!htbp]
    \centering
    \captionsetup[subfigure]{font=small,justification=centering}
    \begin{minipage}[t]{0.485\linewidth}
        \centering
        \textbf{Basic}\par\smallskip
        \begin{subfigure}[t]{0.485\linewidth}
            \centering
            \includegraphics[width=\linewidth]{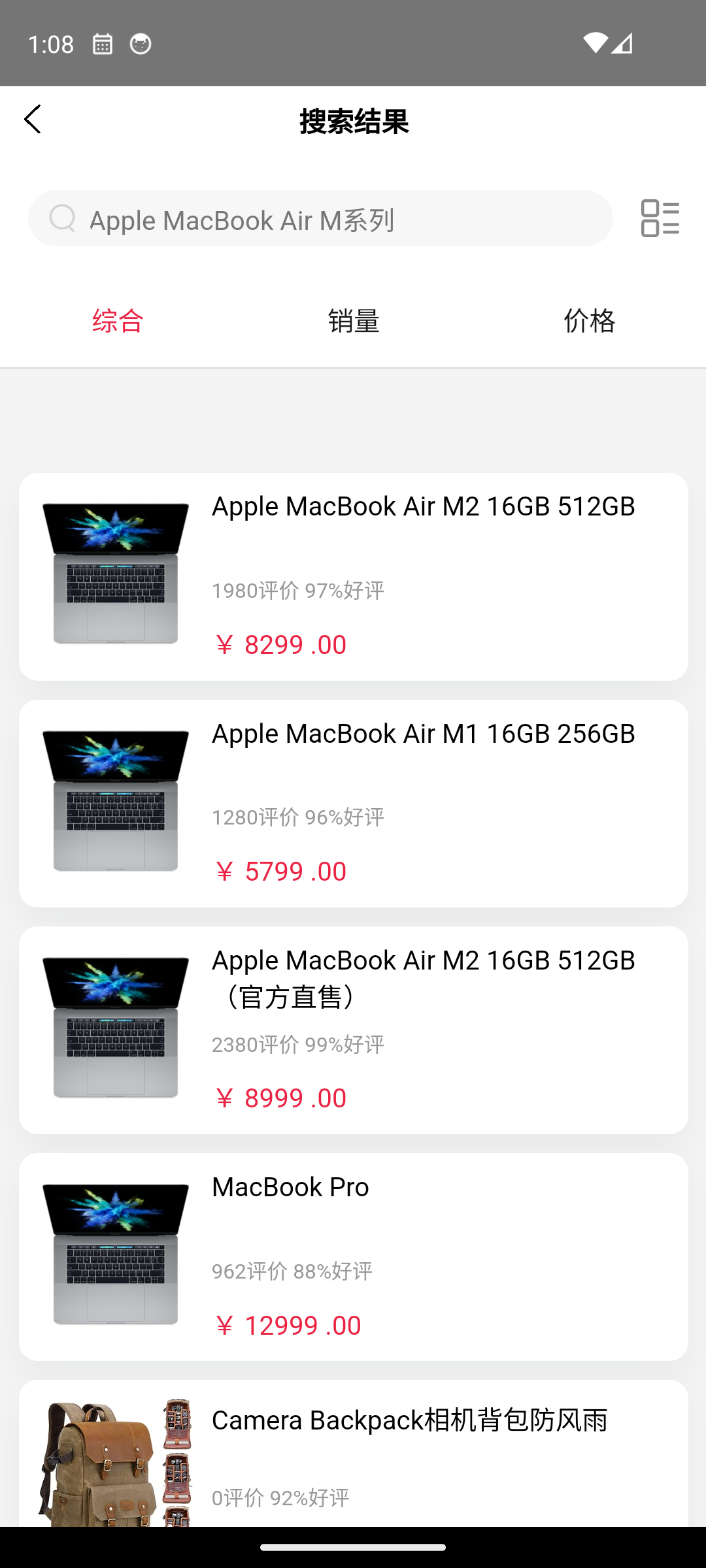}
            \caption{Candidate selection\\Step~9.}
        \end{subfigure}\hfill
        \begin{subfigure}[t]{0.485\linewidth}
            \centering
            \includegraphics[width=\linewidth]{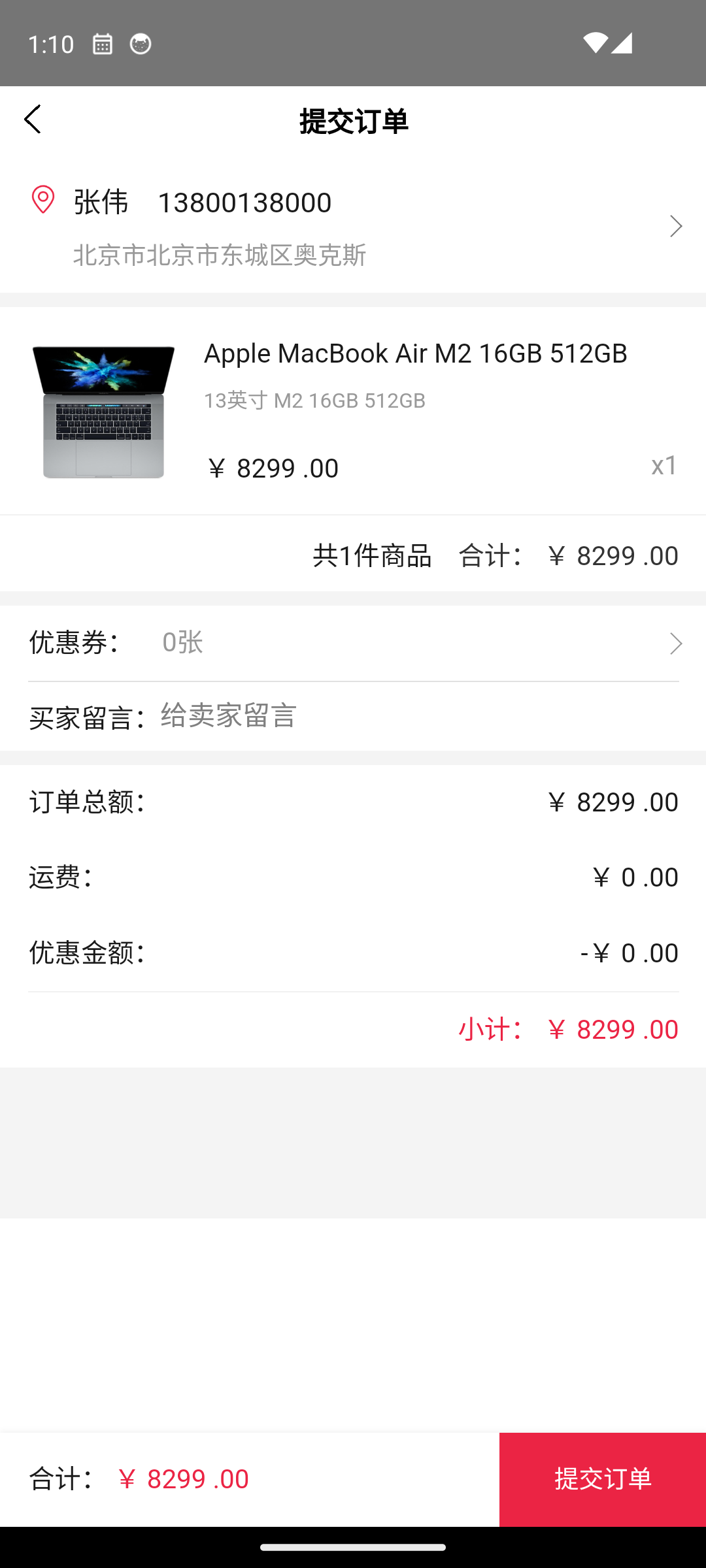}
            \caption{Checkout\\Step~24; CNY~8,299.}
        \end{subfigure}
    \end{minipage}\hfill
    \begin{minipage}[t]{0.485\linewidth}
        \centering
        \textbf{Personalized}\par\smallskip
        \begin{subfigure}[t]{0.485\linewidth}
            \centering
            \includegraphics[width=\linewidth]{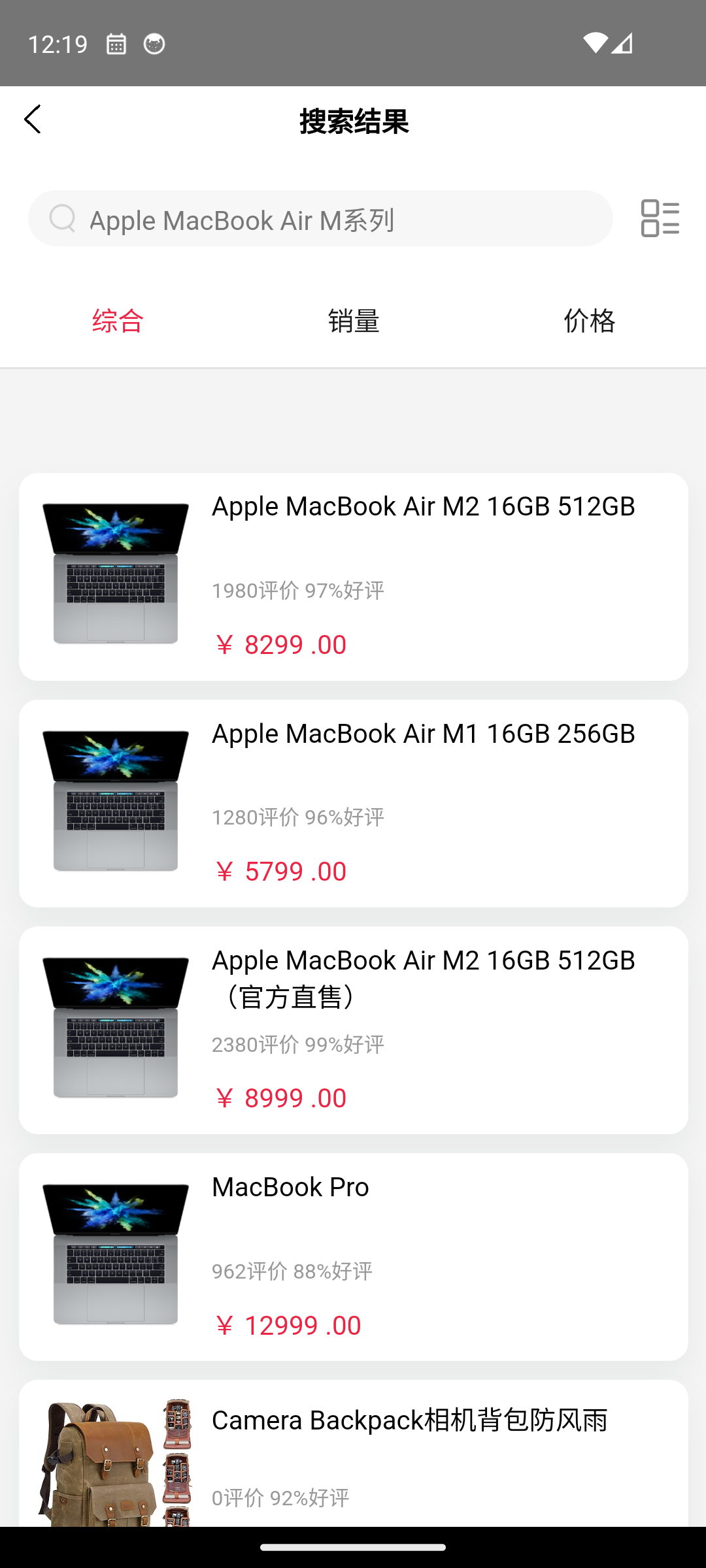}
            \caption{Candidate selection\\Step~8.}
        \end{subfigure}\hfill
        \begin{subfigure}[t]{0.485\linewidth}
            \centering
            \includegraphics[width=\linewidth]{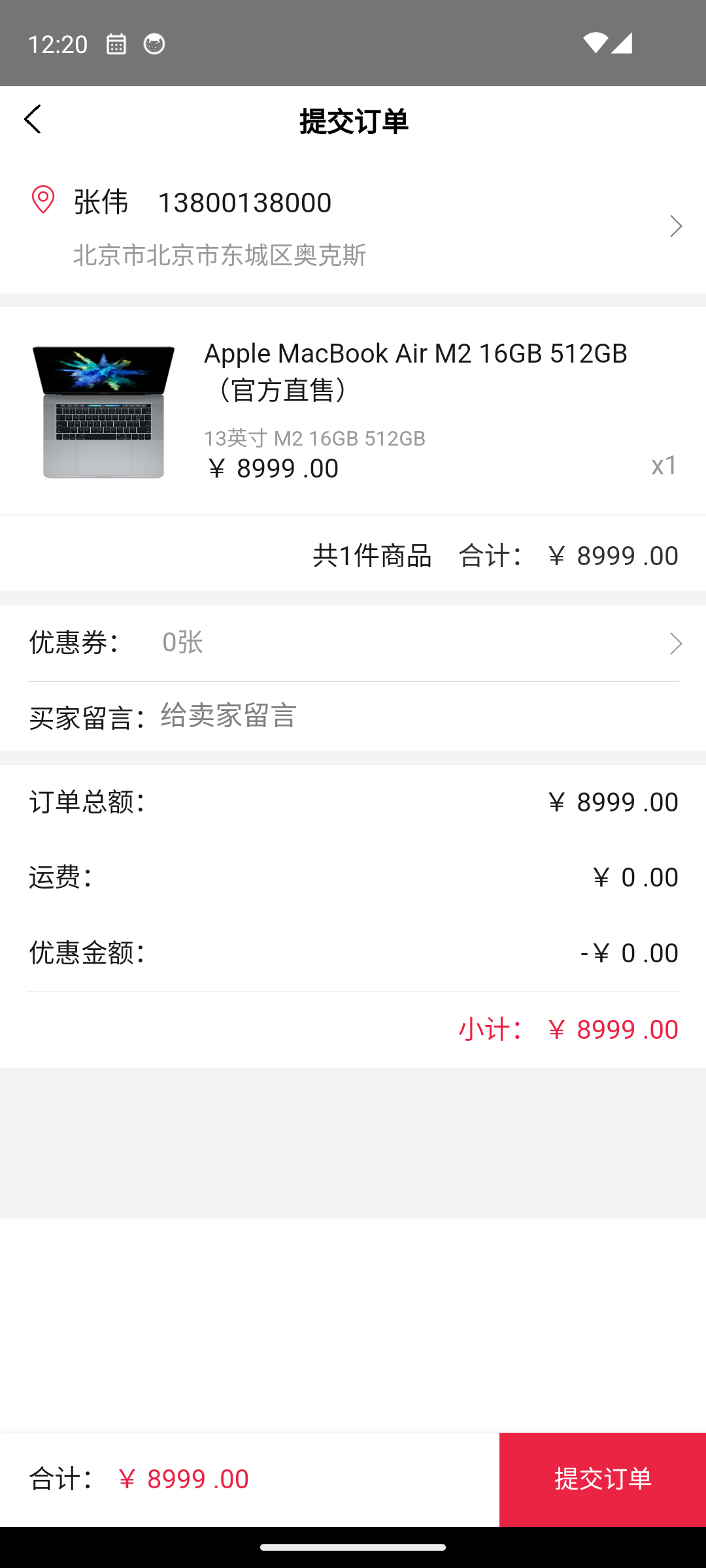}
            \caption{Checkout\\Step~23; CNY~8,999.}
        \end{subfigure}
    \end{minipage}
    \caption{Official-store preference. Both candidate lists include the
    ordinary and official options. Panels (b) and (d) show checkout
    immediately before order submission.}
    \label{fig:appendix-case-laptop}
\end{figure}

Figure~\ref{fig:appendix-case-laptop} shows two available M2 listings
with 16\,GB memory and 512\,GB storage. Basic selects the ordinary-store
listing at CNY~8,299; Personalized selects the official listing at
CNY~8,999. Both runs pass their task-specific checkers. The preference
prioritizes the official source among products that meet the
configuration and price requirements.

\clearpage

\subsection{Staying Within Budget (\textsc{PrefD}~02, Constraint)}
\label{sec:appendix-case-shoes}

\noindent\textbf{Instruction.}
Buy white commuting sneakers in Taodian, select size~42, and submit
the order using the specified delivery details.

\noindent\textbf{Preference.}
Keep the shoe price at or below CNY~300.

\begin{figure}[!htbp]
    \centering
    \captionsetup[subfigure]{font=small,justification=centering}
    \begin{minipage}[t]{0.485\linewidth}
        \centering
        \textbf{Basic}\par\smallskip
        \begin{subfigure}[t]{0.485\linewidth}
            \centering
            \includegraphics[width=\linewidth]{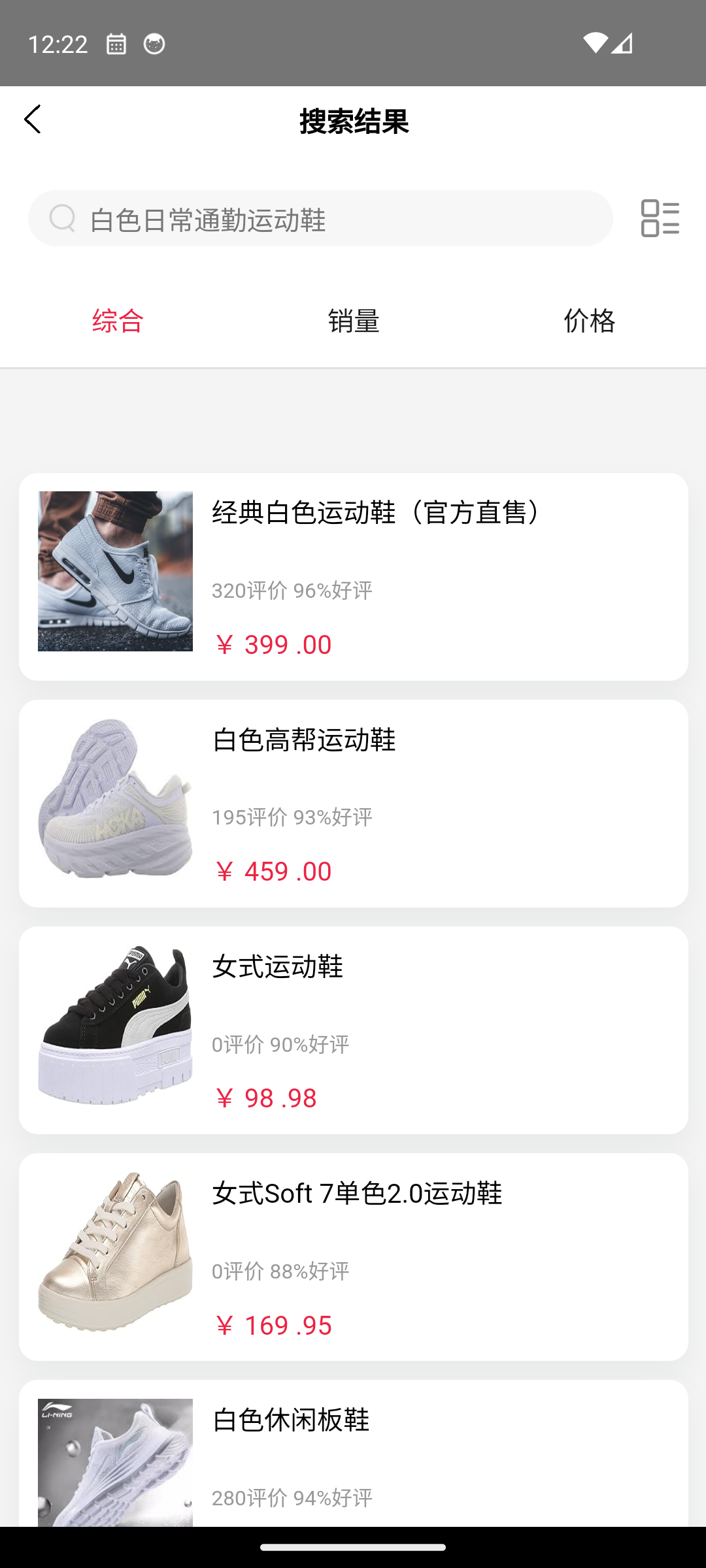}
            \caption{Initial results\\Step~8.}
        \end{subfigure}\hfill
        \begin{subfigure}[t]{0.485\linewidth}
            \centering
            \includegraphics[width=\linewidth]{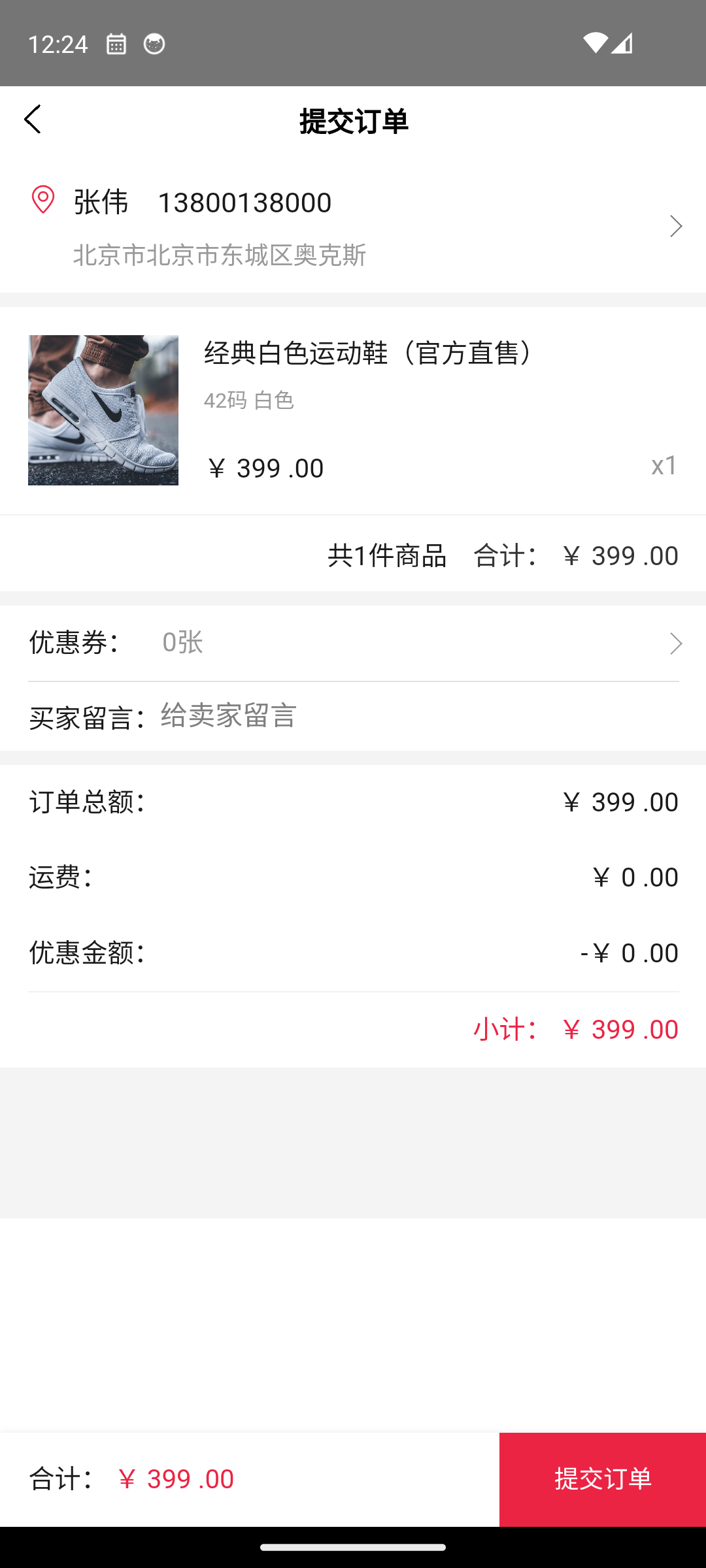}
            \caption{Checkout\\Step~23; CNY~399.}
        \end{subfigure}
    \end{minipage}\hfill
    \begin{minipage}[t]{0.485\linewidth}
        \centering
        \textbf{Personalized}\par\smallskip
        \begin{subfigure}[t]{0.485\linewidth}
            \centering
            \includegraphics[width=\linewidth]{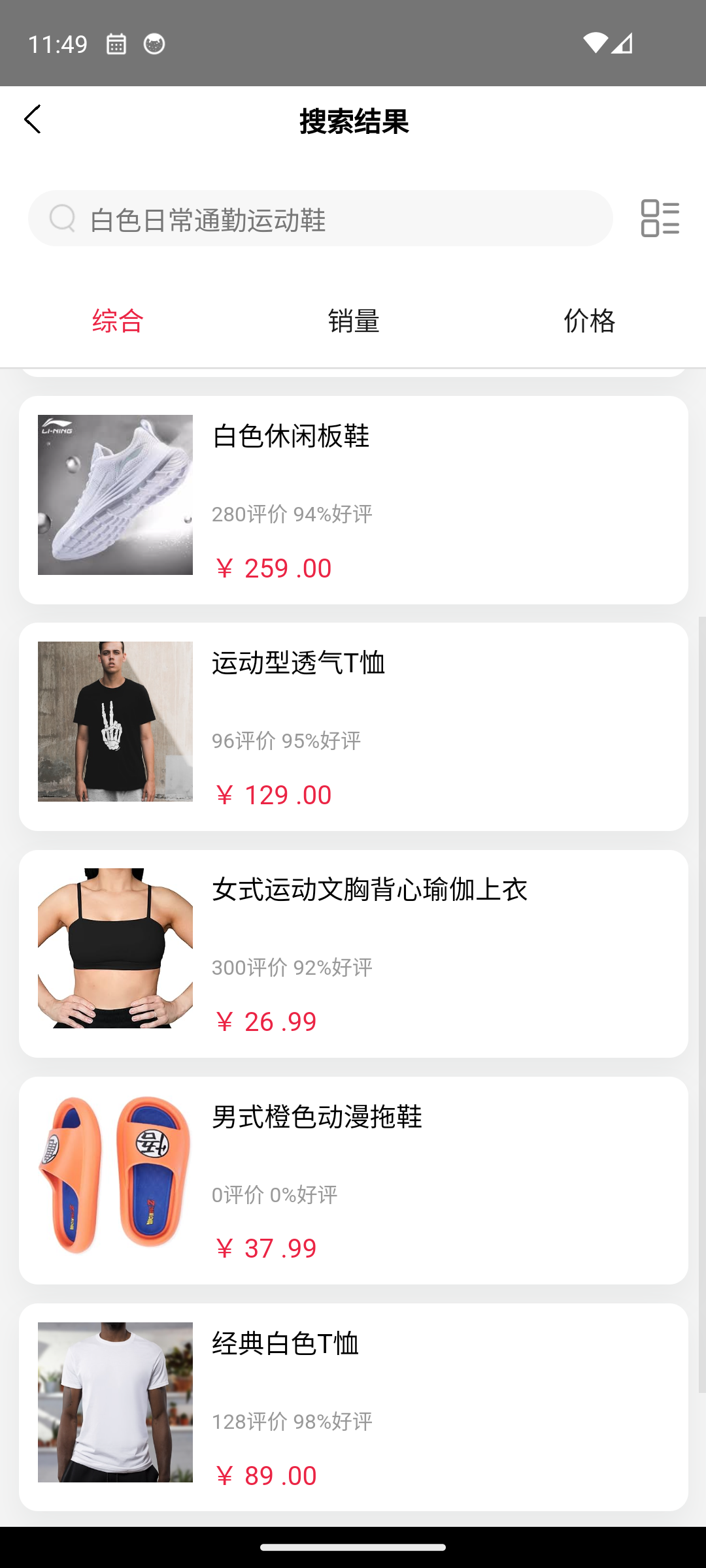}
            \caption{After scrolling\\Step~12.}
        \end{subfigure}\hfill
        \begin{subfigure}[t]{0.485\linewidth}
            \centering
            \includegraphics[width=\linewidth]{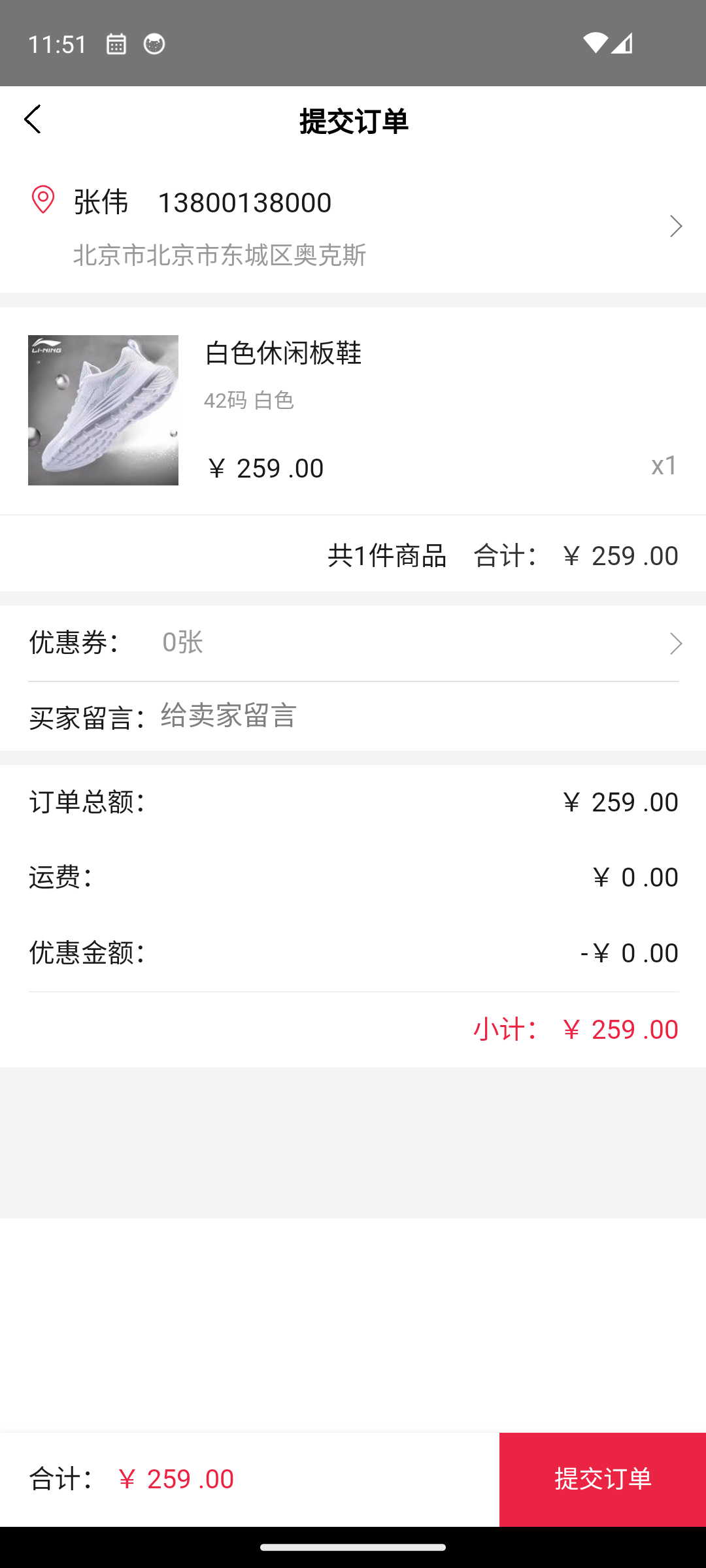}
            \caption{Checkout\\Step~28; CNY~259.}
        \end{subfigure}
    \end{minipage}
    \caption{Budget preference. Further search yields a pair within
    CNY~300. Panels (b) and (d) show the selected white, size-42 shoes
    immediately before order submission.}
    \label{fig:appendix-case-shoes}
\end{figure}

In Figure~\ref{fig:appendix-case-shoes}, Basic selects a CNY~399 pair
from the initial results. Personalized scrolls further and selects
a CNY~259 pair. Both orders contain white, size-42 shoes and pass
their respective checkers. The budget requirement restricts eligible
products and leads to further search before selection.

\end{document}